\documentclass{article}
\usepackage[T1]{fontenc}
\usepackage{microtype}
\usepackage{placeins}
\usepackage{needspace}
\usepackage{graphicx}
\usepackage{booktabs,tabularx,array,multirow}
\usepackage{amsmath,amssymb,mathtools}
\usepackage{algorithm}
\usepackage{algorithmic}
\usepackage{upquote}
\renewcommand{\algorithmiccomment}[1]{\hfill$\triangleright$~#1}
\usepackage[hyperfootnotes=false]{hyperref}
\usepackage{xurl}
\usepackage[preprint]{icml2026}
\usepackage[varqu,scaled=0.98]{zi4}

\newcommand{\bank}{\mathcal{B}}
\newcommand{\policy}{\pi_\theta}
\newcommand{\canon}{\operatorname{canon}}
\newcommand{\CE}{\operatorname{CE}}
\newcommand{\CEPgAuc}{2.3}
\newcommand{\CEPgAucCI}{\mbox{[0.3, 4.3]}}
\newcommand{\CEKeepAucSmall}{97.9}
\newcommand{\CEKeepAccSmall}{98.3}
\newcommand{\ReinforceHeGapSmall}{41.7}
\newcommand{\LikAucSmall}{8.0}
\newcommand{\ReinforceZeroSmall}{98.2}
\newcommand{\LikNearSmall}{66.4}
\newcommand{\LikExtraSmall}{7.2}
\newcommand{\CEKeepAucLarge}{90.2}
\newcommand{\CEKeepAccLarge}{98.9}
\newcommand{\ReinforceHeGapLarge}{43.1}
\newcommand{\LikAucLarge}{3.1}
\newcommand{\ReinforceZeroLarge}{91.5}
\newcommand{\LikNearLarge}{75.0}
\newcommand{\LikExtraLarge}{2.5}
\newcommand{\HeBaseQwenOneSeven}{12.0}
\newcommand{\HeBaseSmolThree}{22.0}
\newcommand{\HeBaseQwenFourteen}{17.1}
\newcommand{\HeBaseQwenThirtyTwo}{21.8}
\newcommand{\SftPairsSmall}{eight}
\newcommand{\SftFoldSmall}{4.5}
\newcommand{\SftFoldSmallCI}{\mbox{[1.8, 7.3]}}
\newcommand{\SftCallsSmall}{88.0}
\newcommand{\SftLowSmall}{82.3}
\newcommand{\SftStackSmall}{8.4}
\newcommand{\SftFoldLarge}{2.1}
\newcommand{\SftFoldLargeCI}{\mbox{[1.3, 12.4]}}
\newcommand{\SftCallsLarge}{192.3}
\newcommand{\SftLowLarge}{91.4}
\newcommand{\SftStackLarge}{9.5}
\newcommand{\HeLowMin}{73.2}
\newcommand{\HeLowMinAbs}{73}
\newcommand{\HeLowMax}{82.2}
\newcommand{\HeLowMaxAbs}{82}
\newcommand{\HeLowGapMin}{37.1}
\newcommand{\HeLowGapMinAbs}{37}
\newcommand{\HeLowGapMax}{58.2}
\newcommand{\HeLowGapMaxAbs}{58}
\newcommand{\LuffyGapMin}{35.8}
\newcommand{\LuffyGapMax}{54.8}
\newcommand{\LuffyGRPOPooled}{6.0}
\newcommand{\LuffyGamma}{0.1}
\newcommand{\LuffyAdv}{$-$0.50}
\newcommand{\LuffyRunsAll}{48}
\newcommand{\LuffyZeroAll}{9}
\newcommand{\LuffyZeroAllCap}{Nine}
\newcommand{\LuffyZeroGRPO}{1}
\newcommand{\LuffyZeroGRPOWord}{one}
\newcommand{\LuffyGPUPairs}{eight}
\newcommand{\ScaleFoldMax}{19}
\newcommand{\ReinforceGapSmall}{28.3}
\newcommand{\ReinforceGapLarge}{32.6}
\newcommand{\PooledAllAUC}{26.9}
\newcommand{\PooledAllAUCCI}{\mbox{[26.2, 27.7]}}
\newcommand{\PooledAllAcc}{38.4}
\newcommand{\PooledAllAccCI}{\mbox{[36.3, 40.6]}}
\newcommand{\SettingsAll}{19}
\newcommand{\ModelsAll}{ten}
\newcommand{\FamiliesAll}{five}
\newcommand{\TasksAll}{five}
\newcommand{\SelfHanoiAcc}{93.6}
\newcommand{\SelfHanoiAccCI}{\mbox{[90.7, 96.0]}}
\newcommand{\SelfHanoiGRPO}{68.3}
\newcommand{\SelfHanoiGRPOCI}{\mbox{[53.2, 80.2]}}
\newcommand{\SelfMatchTotalCalls}{177.6}
\newcommand{\SelfLuffyGapSmall}{56.7}
\newcommand{\SelfLuffyGapLarge}{28.4}
\newcommand{\SelfCurCovSmall}{32}
\newcommand{\SelfCurFnSmall}{40}
\newcommand{\SelfCurCovLarge}{18}
\newcommand{\SelfCurFnLarge}{27}
\newcommand{\SelfStaticPooled}{5.9}
\newcommand{\SelfStaticPooledCI}{\mbox{[2.2, 9.9]}}
\newcommand{\SmolColdTrain}{153}
\newcommand{\SmolColdUnseen}{38}
\newcommand{\SmolColdCovered}{80}
\newcommand{\ColdZeroMbpp}{98.9}
\newcommand{\ColdZeroSmol}{79.3}
\newcommand{\ColdZeroLcb}{96.8}
\newcommand{\ColdBaseMbpp}{0.2}
\newcommand{\ColdBaseLcb}{0.5}
\newcommand{\ColdAllOnlineMbpp}{22.6}
\newcommand{\ColdAllGRPOMbpp}{0.3}
\newcommand{\ColdAllOnlineLcb}{15.8}
\newcommand{\ColdAllGRPOLcb}{1.2}
\newcommand{\ColdCovOnlineMbppAbs}{52}
\newcommand{\ColdCovGRPOMbppAbs}{1}
\newcommand{\MbppLongCovOnlineAbs}{78}
\newcommand{\MbppLongCovGRPOAbs}{7}
\newcommand{\SmolColdEmpty}{73}
\newcommand{\ColdCovOnlineLcbAbs}{47}
\newcommand{\ColdCovGRPOLcbAbs}{4}
\newcommand{\GPUFoldMin}{1.7}
\newcommand{\GPUSelfFold}{6.5}
\newcommand{\GPUMbppFold}{12.2}
\newcommand{\GPUOverhead}{11.7}
\newcommand{\EnvCheckN}{eleven}
\newcommand{\EnvCheckMin}{58.2}
\newcommand{\EnvCheckMax}{96.4}
\newcommand{\ColdEligibleMbpp}{129}
\newcommand{\ColdUnseenMbpp}{26}
\newcommand{\ColdEligibleSmol}{191}
\newcommand{\ColdEligibleLcb}{300}
\newcommand{\ColdUnseenLcb}{60}
\newcommand{\ColdTrainCap}{160}
\newcommand{\ColdUnseenShare}{20}
\newcommand{\OffLeadMin}{25}
\newcommand{\OffLeadMax}{54}
\newcommand{\ProjAucSmall}{2.5}
\newcommand{\ProjAucSmallCI}{\mbox{[0.4, 4.4]}}
\newcommand{\ProjAucSmol}{2.3}
\newcommand{\ProjAucSmolCI}{\mbox{[0.6, 4.0]}}
\newcommand{\ProjAucLarge}{3.2}
\newcommand{\ProjAucLargeCI}{\mbox{[1.5, 4.8]}}
\newcommand{\ProjPooledOnline}{4.4}
\newcommand{\ProjPooledOnlineCI}{\mbox{[1.6, 7.1]}}
\newcommand{\ProjPooledOffline}{5.9}
\newcommand{\ProjPooledOfflineCI}{\mbox{[3.2, 8.5]}}
\newcommand{\CapPromptMbpp}{1,024}
\newcommand{\CapCompletionMbpp}{512}
\newcommand{\CapPromptLcb}{2,048}
\newcommand{\CapCompletionLcb}{768}

\icmltitlerunning{Failed Rollouts Still Teach}
\hypersetup{pdfauthor={Chenyu Zhou, Qiliang Jiang, Shuning Wu, Xu Zhou},pdftitle={Failed Rollouts Still Teach: Teacher-Free Supervision for Code Repair},colorlinks=true,linkcolor=blue,citecolor=blue,urlcolor=blue}
\begin{document}
\raggedbottom
\clubpenalty=10000
\widowpenalty=10000
\displaywidowpenalty=10000
\twocolumn[
\icmltitle{Failed Rollouts Still Teach: Teacher-Free Supervision for Code Repair}
\begin{icmlauthorlist}
\icmlauthor{Chenyu Zhou}{isct}
\icmlauthor{Qiliang Jiang}{zju}
\icmlauthor{Shuning Wu}{nus}
\icmlauthor{Xu Zhou}{nus}
\end{icmlauthorlist}
\icmlaffiliation{isct}{School of Engineering, Institute of Science Tokyo, Tokyo, Japan}
\icmlaffiliation{zju}{College of Control Science and Engineering, Zhejiang University, Hangzhou, China}
\icmlaffiliation{nus}{Department of Electrical and Computer Engineering, National University of Singapore, Singapore}
\icmlcorrespondingauthor{Chenyu Zhou}{zhou.c.76d6@m.isct.ac.jp}
\icmlcorrespondingauthor{Xu Zhou}{zhouxu\_nus@u.nus.edu}
\icmlkeywords{code repair, reinforcement learning, verifiable rewards, sample efficiency, repair witnesses, self-bootstrapping}
\vskip 0.3in
]
\printAffiliationsAndNotice{}
\hypersetup{pdfsubject={Non-anonymous preprint}}
\begin{abstract}
Reinforcement learning for code repair pays one verifier call per proposal, yet a failed call reveals only that the program is wrong, not what it should have been.
On \mbox{HumanEval+} repair, over three quarters of GRPO updates have uniform rewards and no reward gradient.
We introduce repair witnesses, which supervise each failed proposal with the nearest program in a bank of verified implementations under canonical token edit distance.
Bootstrapping the bank from the model's own passing proposals needs no teacher or reference.
One retrieval rule yields three operating points: offline (no online calls), online, and success-projected witnesses.
Under equal online budgets, online witnesses learn faster and finish above GRPO in all \SettingsAll{} settings (\TasksAll{} benchmarks, \ModelsAll{} models, 1.7B to 32B), repairing \HeLowMinAbs\% to \HeLowMaxAbs\% of held-out \mbox{HumanEval+} bugs at low temperature, \HeLowGapMinAbs{} to \HeLowGapMaxAbs~points above GRPO.
With a self-bootstrapped bank, construction included, \mbox{Qwen3-14B} reaches 94\% on a hard repair task in 230 calls and matches GRPO's 1,600-call accuracy of 68\% with \GPUSelfFold~times less training GPU time.
From a cold start on \mbox{MBPP+} and LiveCodeBench, witnesses repair \ColdCovOnlineMbppAbs\% and \ColdCovOnlineLcbAbs\% of held-out bugs on functions that begin with a self-generated program, versus GRPO's \ColdCovGRPOMbppAbs\% and \ColdCovGRPOLcbAbs\% (\MbppLongCovOnlineAbs\% against \MbppLongCovGRPOAbs\% on \mbox{MBPP+} at 1,200 calls).
On every \mbox{HumanEval+} model, online witnesses write correct programs outside the bank more often than GRPO succeeds at all and beat GRPO even when it mixes a reference into every group.
The nearest witness teaches faster than farther ones in every comparison, and the trained policies transfer to unseen functions.
\end{abstract}
\section{Introduction}
\begin{figure*}[t]
\centering
\includegraphics[width=\textwidth]{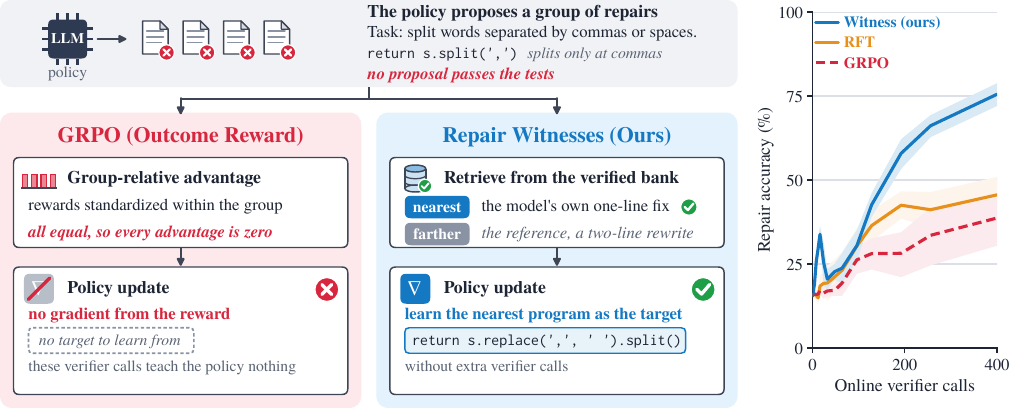}
\caption{Failed proposals still supply a learning target. Left: on a \mbox{HumanEval+} defect in \texttt{words\_string}, every proposal fails, so GRPO gets no reward gradient, while repair witnesses retrieve the nearest verified program in the bank (here the model's own one-line fix, not the two-line reference) as a supervised target. Right: on \mbox{HumanEval+} repair, \mbox{Qwen3-14B} learns faster with repair witnesses than with GRPO or rejection fine-tuning (RFT) at the same verifier budget. Shading shows 95\% seed-bootstrap intervals.}
\label{fig:overview}
\end{figure*}

When reinforcement learning trains a language model to repair code, verifier calls are the unit of cost.
The policy proposes a program, the verifier runs its tests, and the learner receives one bit: pass or fail.
This loop underlies much of the progress in code generation and reasoning \citep{coderl,deepseekmath,deepseekr1}, but on hard repairs most of these bits report failure.
When every proposal in a group receives the same reward, the group-relative advantage is zero and the reward contributes no gradient at all.
On \mbox{HumanEval+} repair, 78\% to 98\% of group relative policy optimization (GRPO) updates carry all-zero advantages.
On Hanoi-Hard, the budget runs out before the policy improves: \mbox{Qwen3-14B} with GRPO ends at 41.9\% after 400 calls, no higher than its 43.0\% base accuracy.

Existing remedies leave this problem open.
Rejection fine-tuning and self-training keep only verified successes and discard the failures \citep{star,raft,restem}.
Feedback methods return explanations or execution traces at the cost of further calls or turns \citep{selfdebug,rlef}.
Methods mixing expert solutions into on-policy updates imitate a fixed reference regardless of what the policy wrote \citep{luffy,relift,chord}, and rewriting a failed attempt takes a stronger model \citep{projdistill}.

A failed call tells the policy only that its program is wrong, leaving no target to move toward.
Yet correct programs already exist, and the model writes them itself: a policy that fails on one defect passes the verifier on other samples for the same function.
Each such program is a verified implementation bought with one verifier call.
The one nearest to a failed attempt changes the least of it.
Verification decides whether a repair is correct.
Distance to verified programs decides what a failure should teach.

We introduce repair witnesses, which turn failed verifier calls into supervision (Figure~\ref{fig:overview}).
A bank holds verified implementations of each training function, and the model's own passing proposals suffice to build it.
A failed proposal retrieves its repair witness, the bank member nearest to it in canonical token edit distance, and learns from that program without extra verifier calls.
One retrieval rule defines three operating points.
Offline witnesses supervise each training defect with its nearest bank member, without online calls.
Online witnesses, our default, retrieve a witness for each failure as the policy learns, keep successful proposals as their own targets, and add a group-centered policy gradient.
Success-projected witnesses also project successful proposals onto the bank.

We evaluate on \TasksAll{} benchmarks (\mbox{HumanEval+} repair of human-written bugs \citep{octopack,evalplus}, Hanoi-Hard, Algo-8, and cold-start \mbox{MBPP+} and LiveCodeBench) and test transfer to unseen functions.
With a bank of only \mbox{Qwen3-14B}'s own proposals, repair witnesses reach 93.6\% on Hanoi-Hard in 230 verifier calls (construction included), whereas GRPO reaches 68.3\% in 1,600 calls (Figure~\ref{fig:scale}).
On functions that begin with a self-generated program, self-bootstrapped \mbox{HumanEval+} banks outperform curated banks with the EvalPlus reference, and from a cold start on \mbox{MBPP+} and LiveCodeBench, online witnesses repair about half of the held-out bugs on such functions while GRPO barely moves (Figure~\ref{fig:cold}, Table~\ref{tab:selfcold}).
Our contributions are as follows.
\begin{itemize}\setlength{\itemsep}{1pt}\setlength{\parskip}{0pt}
\item \textbf{Self-bootstrapped repair witnesses.} Each failure learns from its nearest program in a bank of verified implementations without extra calls, and banks of the model's own passing proposals need no teacher or reference solution. One retrieval rule gives three operating points: offline witnesses beat all outcome-reward and reference-mixing baselines without online calls, online witnesses add accuracy and outside-bank correctness, success-projected witnesses are more accurate still, and offline and online witnesses compose (\mbox{Sections}~\ref{sec:method}, \ref{sec:selfbank}, and~\ref{sec:family}).
\item \textbf{Call, GPU-time, and scale efficiency.} Online witnesses beat GRPO in all \SettingsAll{} settings across \ModelsAll{} models from \FamiliesAll{} families, at every GRPO learning rate and batch shape we tested, and from a shared offline-witness start. They reach GRPO's accuracy with less training GPU time in every timed comparison, lift 1.7B and 3B models above 14B to 32B GRPO models, stay ahead when GRPO gets four times the budget, and learn from a cold start where GRPO, DAPO, and Dr.~GRPO stall (\mbox{Sections}~\ref{sec:he} to~\ref{sec:selfbank}).
\item \textbf{Why the model's own nearest fix.} Learning comes from the witness targets: without them, REINFORCE only matches GRPO, and they alone keep most of the lead of online witnesses. The nearest witness teaches faster than farther ones in every comparison, and the model's own programs, more likely under the policy, beat an external reference placed in a curated bank or added to every GRPO group. Witness-trained policies write correct programs outside the bank more often than GRPO succeeds at all and transfer to unseen functions (\mbox{Sections}~\ref{sec:he}, \ref{sec:selfbank}, and~\ref{sec:mechanism}).
\end{itemize}

\section{Self-Bootstrapped Repair Witnesses}
\label{sec:method}
\subsection{Code Repair under a Verifier Budget}
An input $x$ contains a function specification and a defective implementation, and the policy $\policy(y\mid x)$ outputs a complete repaired function $y$.
The verifier $V_f(y)\in\{0,1\}$ runs the tests of function $f$ on $y$.
We count each verification as one verifier call, whether the program passes, fails, or times out.
The online budget of a run counts the calls on proposals (rollouts) sampled during training, and the total budget adds every call spent building the bank.
Each training function has a bank $\bank_f$ of verified implementations.
Held-out defects differ from training defects and bank programs, and the policy repairs them without the bank.

\subsection{Self-Bootstrapped Banks}
\label{sec:selfbank-method}
A self-bootstrapped bank holds only the policy's own passing programs, with no teacher or reference implementation.
Before the first update, we sample repair proposals or specification-conditioned generations from the base model, verify each once, and keep the passing, canonically distinct ones as the bank (Appendices~\ref{app:budget} and~\ref{app:canonical}).
A growing bank admits each canonically new passing proposal after the update that sampled it, without extra calls.
A curated bank combines a reference implementation with verified alternatives and stays fixed.

\subsection{Nearest Repair Witnesses}
For a program $q$, the repair witness is
\[
 w^\star_f(q)=\arg\min_{b\in\bank_f}
 d\bigl(\canon(q),\canon(b)\bigr),
\]
where $d$ is token Levenshtein distance and $\canon$ strips comments and formatting and renames local variables (Appendix~\ref{app:canonical}).
Ties are broken uniformly at random.
Because distance is measured from each new proposal, the witness follows the policy as it learns, and failures on the same input can retrieve different bank members.
Retrieval calls neither a model nor the verifier.

\subsection{A Family of Witness Objectives}
\label{sec:family-method}
The retrieval rule defines a family of three objectives.

\textbf{Online witnesses} combine a group-centered policy gradient \citep{reinforce} with supervision on verified targets.
Each batch samples $G$ proposals for each of $m$ inputs from the current policy.
For proposal $y_i$ of input $x_i$ and function $f_i$, let $r_i=V_{f_i}(y_i)$ and let $\widehat A_i$ be the reward centered within its input's group.
Write $\bar\ell_\theta(z\mid x)$ for the mean continuation-token log probability of $z$, including the end-of-sequence token when present.
For a batch of $n=mG$ proposals, the loss is
\begin{equation}
\begin{split}
\mathcal{L}(\theta)
={}&-\frac{1}{n}\sum_{i=1}^{n}
\widehat A_i\,\bar\ell_\theta(y_i\mid x_i)\\
&+\frac{\lambda}{n}\sum_{i=1}^{n}\CE_\theta(t_i\mid x_i),\\[-2pt]
t_i={}&\begin{cases}
y_i,&r_i=1,\\
w^\star_{f_i}(y_i),&r_i=0,\ \bank_{f_i}\neq\varnothing,\\
\varnothing,&r_i=0,\ \bank_{f_i}=\varnothing,
\end{cases}
\end{split}
\label{eq:loss}
\end{equation}
where $\CE_\theta=-\bar\ell_\theta$, $\CE_\theta(\varnothing\mid x)=0$, $\lambda=1$, and advantages are treated as constants under differentiation.
When a whole group fails, the policy-gradient term vanishes but the supervised term remains wherever the bank holds a member.
Algorithm~\ref{alg:update} in Appendix~\ref{app:canonical} lists one update.

\textbf{Offline witnesses} keep only the supervised term and run without online calls: each training input is supervised with $t_i=w^\star_{f_i}(\tilde x_i)$, the witness of its defective program $\tilde x_i$.

\textbf{Success-projected witnesses} set $t_i=w^\star_{f_i}(y_i)$ for every proposal, passing or failing, and keep the policy-gradient term.

As baselines, rejection fine-tuning keeps only success supervision and REINFORCE only the policy gradient.

\section{Experimental Setup}
\label{sec:setup}

\subsection{Tasks, Models, and Banks}
\mbox{HumanEval+} repair asks the policy to fix the human-written bugs of HumanEvalFix \citep{octopack} on HumanEval specifications \citep{codex}, and the extended EvalPlus tests verify each repair \citep{evalplus}.
For each model and function, we sample 16 programs from the specification and verify each once.
Failures among these samples become the natural training defects, and passes join the EvalPlus reference in the bank.
Most bank members are the model's own programs (Table~\ref{tab:data} in Appendix~\ref{app:data}), and the bank excludes the original implementation behind each human-written bug.
For each model, a fixed rule selects functions with base repair accuracy at most 75\%, and all methods share that set.
We train \mbox{Qwen3-1.7B}, \mbox{Qwen3-14B}, \mbox{Qwen3-32B} \citep{qwen3}, and \mbox{SmolLM3-3B} \citep{smollm3}, and evaluate each on the held-out bugs (Appendix~\ref{app:data}).
For \mbox{Qwen3-1.7B} and \mbox{Qwen3-14B} we also train on self-bootstrapped banks, which drop every EvalPlus reference and grow with passing proposals.

Hanoi-Hard gives every model the same 40 training and 40 held-out defects of cyclic and linear Tower-of-Hanoi variants, each produced by two single-node abstract-syntax-tree mutations of a verified implementation.
We train eight models on it: \mbox{Qwen3-1.7B}, \mbox{SmolLM3-3B}, \mbox{Granite-4.2-3B} \citep{granite42}, \mbox{Qwen3-14B}, \mbox{Mistral-Small-3.2-24B} (\mbox{Mistral-24B}; \citealp{mistralsmall32}), \mbox{Qwen3.6-27B} \citep{qwen36}, \mbox{Muse Glimmer-30B} \citep{museglimmer}, and \mbox{Qwen3-32B}.
All eight share a curated bank of six implementations per function.
For \mbox{Qwen3-14B} we also build a self-bootstrapped bank from verified base-model proposals (Appendix~\ref{app:budget}).

Algo-8 uses eight QuixBugs functions \citep{quixbugs}, each with 20 training and 20 held-out single-node mutation defects and six curated bank implementations.
We train \mbox{Qwen3-1.7B}, \mbox{SmolLM3-3B}, \mbox{Granite-4.2-3B}, and \mbox{Qwen3.5-4B} \citep{qwen35} on it.

We test a cold start with \mbox{Qwen3-1.7B} and \mbox{SmolLM3-3B} on \mbox{MBPP+} \citep{mbpp,evalplus} and \mbox{Qwen3-8B} on LiveCodeBench \citep{livecodebench}.
Each benchmark uses only the model's own specification-conditioned samples: failures become the defects and passes form a growing self-bootstrapped bank.
Initial banks cover 39 of 103 training functions on \mbox{MBPP+} with \mbox{Qwen3-1.7B}, \SmolColdCovered{} of \SmolColdTrain{} with \mbox{SmolLM3-3B}, and 51 of 160 on LiveCodeBench, with further functions held out as unseen (Appendix~\ref{app:data}).

We evaluate transfer without further training on 69 HumanEvalFix and 10 QuixBugs functions absent from the training sets and banks (Appendix~\ref{app:transfer}).

\subsection{Methods and Budgets}
\label{sec:budgets}
Online methods default to 400 verifier calls and draw a fresh batch for every update.
Online witnesses take 100 updates of two inputs with two proposals each and use one optimization configuration for every model and task.
GRPO \citep{deepseekmath} takes 100 updates of four proposals for one input, normalizes centered rewards by the group standard deviation, and adds a KL penalty toward the base policy.
On \mbox{HumanEval+} (Table~\ref{tab:he} and Table~\ref{tab:variants} in Appendix~\ref{app:training}), further outcome-reward baselines are rejection fine-tuning (RFT), larger-group GRPO ($G=8$), KL-free GRPO, decoupled clip and dynamic sampling policy optimization (DAPO), and Dr.~GRPO \citep{dapo,drgrpo}, with the last two also on \mbox{MBPP+}. LUFFY-style off-policy guidance \citep{luffy} adds the reference to every GRPO group.
Algo-8 and \mbox{HumanEval+} add group-centered REINFORCE.
We also give GRPO four times the budget on Hanoi-Hard and continue GRPO and online witnesses to 1,200 calls on \mbox{MBPP+}.

Offline witnesses match online witnesses in inputs, batch size, and update count, and success-projected witnesses keep the online configuration.
Unless stated otherwise, repair witnesses and witness-trained policies refer to the online form.
The random farther control keeps the bank and loss fixed and, when a predesignated far anchor lies beyond the nearest witness, swaps that witness for a random strictly farther member (Appendix~\ref{app:canonical}).

\subsection{Metrics}
\label{sec:metrics}
We measure learning efficiency by the normalized area under the 400-call held-out accuracy curve (AUC) and by accuracy at the end of training.
Accuracy is the pass rate at the training temperature $T=1$, and we also decode final policies at $T=0.2$.
Rates are percentages, and all differences, including AUC differences, are in percentage points (points).
Fewer calls divides a baseline's budget by the call count after which witnesses stay at or above its final accuracy, including bank-construction calls for self-bootstrapped banks.
Less GPU time divides the baseline's full-budget training GPU time by the witnesses' time up to that call count (Appendix~\ref{app:statistics}).
Where the 95\% interval of calls to match the baseline reaches zero, the ratio has no upper bound and its lower 95\% bound $L$ is shown as $>L$. A dash marks an untimed or inapplicable entry.
Outside-bank correctness is the fraction of all outputs that pass and differ canonically from every bank member.
We pair methods on the same training seeds, report 95\% paired seed-bootstrap intervals, and weight settings equally in pooled effects (Appendix~\ref{app:statistics}).

\section{Results}
\label{sec:results}
\subsection{Faster Learning at the Same Verifier Budget}
\label{sec:he}
\begin{table*}[!t]
\caption{Repair witnesses outperform GRPO, RFT, and LUFFY-style guidance on \mbox{HumanEval+} repair, and offline witnesses do so without online calls. At $T=1$, Outside bank and $\Delta$ (online witness minus the row's method) follow Section~\ref{sec:metrics}, as do Fewer calls, Less GPU time, $>$, dashes, and the 95\% paired seed-bootstrap intervals in brackets. Bold marks each model's highest $T=1$ accuracy. Before training, the base models repair \HeBaseQwenOneSeven\%, \HeBaseSmolThree\%, \HeBaseQwenFourteen\%, and \HeBaseQwenThirtyTwo\% of held-out bugs over 16 attempts each.}
\label{tab:he}
\centering\small\setlength{\tabcolsep}{3.5pt}
\begin{tabular*}{\linewidth}{@{\extracolsep{\fill}}llrrrrr@{\extracolsep{0pt}\hspace{3pt}}>{\scriptsize}l@{\extracolsep{\fill}\hspace{2\tabcolsep}}r@{\extracolsep{0pt}\hspace{3pt}}>{\scriptsize}l@{\extracolsep{\fill}\hspace{2\tabcolsep}}r@{\extracolsep{0pt}\hspace{3pt}}>{\scriptsize}l@{\extracolsep{\fill}\hspace{2\tabcolsep}}r@{\extracolsep{0pt}\hspace{3pt}}>{\scriptsize}l@{}}
\toprule
 &  &  & \multicolumn{2}{c}{Accuracy (\%)} & Outside & \multicolumn{2}{c}{$\Delta$AUC} & \multicolumn{2}{c}{$\Delta$ accuracy} & \multicolumn{2}{c}{Fewer} & \multicolumn{2}{c}{Less GPU} \\
\cmidrule(lr){4-5}
Model & Method & Calls & $T{=}1$ & $T{=}0.2$ & bank (\%) & \multicolumn{2}{c}{[95\% CI]} & \multicolumn{2}{c}{[95\% CI]} & \multicolumn{2}{c}{calls ($\times$)} & \multicolumn{2}{c}{time ($\times$)} \\
\midrule
\mbox{Qwen3-1.7B} & Online witness & 400 & \textbf{58.6} & 73.2 & 40.2 & -- &  & -- &  & -- &  & -- &  \\
 & Offline witness & 0 & 57.4 & 70.6 & 30.4 & -- &  & -- &  & -- &  & -- &  \\
 & GRPO & 400 & 15.6 & 15.0 & 15.2 & +17.6 & \mbox{[14.8, 20.4]} & +43.0 & \mbox{[32.7, 51.0]} & 3.3 & \mbox{[2.7, 3.7]} & 1.8 & \mbox{[1.6, 2.1]} \\
 & RFT & 400 & 24.7 & 24.6 & 23.9 & +14.1 & \mbox{[8.5, 19.1]} & +33.9 & \mbox{[22.8, 44.5]} & 2.6 & \mbox{[1.8, 3.4]} & 1.3 & \mbox{[1.0, 1.6]} \\
 & LUFFY-style & 400 & 3.8 & 9.1 & 2.9 & +24.1 & \mbox{[21.2, 27.0]} & +54.8 & \mbox{[50.8, 58.6]} & 19.1 & \mbox{[4.6, 293.9]} & $>$4.8 &  \\
\addlinespace[3pt]
\mbox{SmolLM3-3B} & Online witness & 400 & \textbf{66.1} & 81.5 & 40.7 & -- &  & -- &  & -- &  & -- &  \\
 & Offline witness & 0 & 64.1 & 78.6 & 21.7 & -- &  & -- &  & -- &  & -- &  \\
 & GRPO & 400 & 26.2 & 37.3 & 24.8 & +15.4 & \mbox{[12.4, 18.3]} & +39.9 & \mbox{[31.4, 48.3]} & 3.4 & \mbox{[3.0, 4.4]} & 3.2 & \mbox{[2.8, 4.0]} \\
 & RFT & 400 & 34.9 & 44.2 & 33.7 & +11.1 & \mbox{[8.0, 13.9]} & +31.2 & \mbox{[24.3, 38.0]} & 2.7 & \mbox{[2.0, 3.1]} & 1.9 & \mbox{[1.5, 2.1]} \\
 & LUFFY-style & 400 & 21.0 & 41.4 & 19.0 & +25.4 & \mbox{[23.0, 27.7]} & +45.1 & \mbox{[38.8, 51.4]} & 4.1 & \mbox{[3.5, 4.7]} & 4.7 & \mbox{[4.0, 5.4]} \\
\addlinespace[3pt]
\mbox{Qwen3-14B} & Online witness & 400 & \textbf{75.5} & 81.9 & 53.5 & -- &  & -- &  & -- &  & -- &  \\
 & Offline witness & 0 & 74.4 & 82.5 & 35.8 & -- &  & -- &  & -- &  & -- &  \\
 & GRPO & 400 & 38.7 & 38.6 & 34.6 & +23.2 & \mbox{[18.7, 28.4]} & +36.8 & \mbox{[28.9, 45.6]} & 3.4 & \mbox{[2.7, 4.3]} & 2.3 & \mbox{[1.8, 3.2]} \\
 & RFT & 400 & 45.6 & 47.1 & 42.7 & +15.9 & \mbox{[11.2, 21.1]} & +29.9 & \mbox{[23.0, 37.6]} & 2.7 & \mbox{[2.3, 3.4]} & 1.6 & \mbox{[1.4, 1.8]} \\
 & LUFFY-style & 400 & 37.1 & 44.5 & 34.6 & +15.8 & \mbox{[11.0, 20.9]} & +38.4 & \mbox{[27.4, 50.8]} & 3.5 & \mbox{[2.5, 5.1]} & 3.2 & \mbox{[2.3, 4.8]} \\
\addlinespace[3pt]
\mbox{Qwen3-32B} & Online witness & 400 & \textbf{74.2} & 82.2 & 54.7 & -- &  & -- &  & -- &  & -- &  \\
 & Offline witness & 0 & 69.1 & 76.5 & 28.4 & -- &  & -- &  & -- &  & -- &  \\
 & GRPO & 400 & 43.7 & 45.1 & 41.7 & +17.8 & \mbox{[14.1, 21.3]} & +30.5 & \mbox{[25.3, 36.2]} & 2.8 & \mbox{[2.2, 3.7]} & 1.7 & \mbox{[1.4, 2.3]} \\
 & LUFFY-style & 400 & 38.4 & 44.4 & 35.3 & +16.1 & \mbox{[11.7, 20.7]} & +35.8 & \mbox{[28.4, 44.5]} & 3.5 & \mbox{[2.4, 4.9]} & 2.5 & \mbox{[1.8, 3.5]} \\
\bottomrule
\end{tabular*}

\end{table*}
\begin{table*}[!t]
\caption{Repair witnesses outperform GRPO on Hanoi-Hard for all eight models. Accuracies are percentages: Witness and GRPO at 400 calls, and Base over 16 samples per defect before training. Differences follow Table~\ref{tab:he}, and the two ratios and $>$ follow Section~\ref{sec:metrics}. Less GPU time marked $^\dagger$ is estimated from total run time (Appendix~\ref{app:statistics}).}
\label{tab:hanoi}
\centering\small\setlength{\tabcolsep}{4.5pt}
\begin{tabular*}{\linewidth}{@{\extracolsep{\fill}}lrrrr@{\extracolsep{0pt}\hspace{3pt}}>{\scriptsize}l@{\extracolsep{\fill}\hspace{2\tabcolsep}}r@{\extracolsep{0pt}\hspace{3pt}}>{\scriptsize}l@{\extracolsep{\fill}\hspace{2\tabcolsep}}r@{\extracolsep{0pt}\hspace{3pt}}>{\scriptsize}l@{\extracolsep{\fill}\hspace{2\tabcolsep}}r@{\extracolsep{0pt}\hspace{3pt}}>{\scriptsize}l@{}}
\toprule
Model & Base & Witness & GRPO & \multicolumn{2}{c}{$\Delta$AUC [95\% CI]} & \multicolumn{2}{c}{$\Delta$ accuracy [95\% CI]} & \multicolumn{2}{c}{Fewer calls ($\times$)} & \multicolumn{2}{c}{Less GPU time ($\times$)} \\
\midrule
\mbox{Qwen3-1.7B} & 7.2 & 77.0 & 8.5 & +45.7 & \mbox{[43.9, 47.5]} & +68.5 & \mbox{[63.8, 73.0]} & 6.5 & \mbox{[5.7, 8.4]} & 4.0 & \mbox{[3.5, 5.3]} \\
\mbox{SmolLM3-3B} & 5.5 & 79.2 & 9.5 & +56.9 & \mbox{[54.8, 59.0]} & +69.7 & \mbox{[65.0, 74.2]} & 15.4 & \mbox{[12.8, 24.3]} & 12.0 & \mbox{[10.3, 17.8]} \\
\mbox{Granite-4.2-3B} & 13.0 & 80.0 & 1.3 & +62.5 & \mbox{[59.8, 65.2]} & +78.7 & \mbox{[73.4, 83.5]} & $>$45 &  & $>$32 &  \\
\mbox{Qwen3-14B} & 43.0 & 87.8 & 41.9 & +27.8 & \mbox{[23.0, 32.6]} & +45.9 & \mbox{[34.7, 57.4]} & 5.4 & \mbox{[4.7, 6.5]} & 2.9 & \mbox{[2.5, 3.5]}\rlap{$^\dagger$} \\
\mbox{Mistral-24B} & 10.5 & 77.1 & 3.4 & +65.6 & \mbox{[61.5, 70.1]} & +73.7 & \mbox{[56.2, 88.3]} & $>$66 &  & $>$60 &  \\
\mbox{Qwen3.6-27B} & 45.0 & 90.9 & 67.1 & +19.3 & \mbox{[14.8, 23.9]} & +23.8 & \mbox{[8.4, 41.9]} & 11.1 & \mbox{[2.9, 26.6]} & 5.8 & \mbox{[2.0, 12.0]}\rlap{$^\dagger$} \\
\mbox{Muse Glimmer-30B} & 13.1 & 78.5 & 18.8 & +56.1 & \mbox{[50.7, 60.8]} & +59.7 & \mbox{[49.0, 70.0]} & $>$21 &  & $>$18 &  \\
\mbox{Qwen3-32B} & 36.6 & 82.9 & 53.4 & +20.6 & \mbox{[18.2, 22.9]} & +29.5 & \mbox{[20.4, 37.2]} & 5.7 & \mbox{[3.8, 6.6]} & 3.3 & \mbox{[2.4, 3.8]}\rlap{$^\dagger$} \\
\bottomrule
\end{tabular*}

\end{table*}

On \mbox{HumanEval+} repair, online witnesses beat GRPO, RFT, and LUFFY-style guidance in AUC and final accuracy and nearly double GRPO's accuracy on \mbox{Qwen3-14B} (Table~\ref{tab:he}).
They match the final accuracy of these baselines and four GRPO variants with at least 2.6~times fewer calls (Table~\ref{tab:he} and Table~\ref{tab:variants} in Appendix~\ref{app:training}).
At $T=0.2$ the lead widens: witness-trained policies repair \HeLowMin\% to \HeLowMax\% of the held-out bugs, \HeLowGapMin{} to \HeLowGapMax~points above GRPO.

GRPO spends most of its budget on updates whose reward teaches nothing: all proposals in a group receive the same reward, even at $G=8$ (Table~\ref{tab:variants}).
DAPO skips such groups, which make up 99.1\% of those it samples on \mbox{Qwen3-1.7B}.
Every witness update carries a verified target for each proposal whose function has a bank member.
Tuning does not rescue GRPO: no tested learning rate ($3\times10^{-5}$, $3\times10^{-4}$) or batch shape (two inputs of four proposals over 50 updates) beats standard GRPO on \mbox{Qwen3-1.7B} or \mbox{Qwen3-14B}, and online witnesses lead in every setting (Table~\ref{tab:variants}).

Giving GRPO verified programs does not close the gap, whether they are mixed into every group or train the policy it starts from.
With a reference solution in every group, LUFFY-style guidance trails online witnesses by \LuffyGapMin{} to \LuffyGapMax~points and finishes \LuffyGRPOPooled~points below GRPO, pooled over the four models (Table~\ref{tab:he}).
\LuffyZeroAllCap{} of its \LuffyRunsAll{} runs collapse to zero accuracy, against \LuffyZeroGRPOWord{} GRPO run and no online witness run (Appendix~\ref{app:training}).
From a shared offline-witness start, online witnesses lead GRPO in AUC and in final accuracy at both temperatures on \mbox{Qwen3-1.7B} and \mbox{Qwen3-14B} and reach its 400-call accuracy with \SftFoldSmall{} and \SftFoldLarge~times fewer calls (Table~\ref{tab:sftrl} in Appendix~\ref{app:copy}).

Counting only correct programs outside the bank, online witnesses exceed the total accuracy of every outcome-reward and reference-mixing baseline, beating GRPO's by 11.0 to 24.6~points on all four models (Tables~\ref{tab:he} and~\ref{tab:variants}).

\subsection{Witnesses Teach Where GRPO Stalls}
\label{sec:scale}

On Hanoi-Hard, repair witnesses outperform GRPO in AUC and final accuracy on all eight models and reach GRPO's 400-call accuracy with at least 5.4~times fewer calls (Table~\ref{tab:hanoi}).
The lead is largest where base accuracy is low and outcome rewards have little to reinforce.
Supervising failures also lets small models overtake large ones: all three witness-trained models at 1.7B and 3B finish above every GRPO-trained model from 14B to 32B, up to \ScaleFoldMax~times their size (Table~\ref{tab:hanoi}).

Including \mbox{Qwen3.5-4B} on Algo-8 (Table~\ref{tab:algo} in Appendix~\ref{app:effects}) and \mbox{Qwen3-8B} on LiveCodeBench, we test \ModelsAll{} models from \FamiliesAll{} families.
Repair witnesses lead GRPO in AUC and final accuracy across all \SettingsAll{} model-task settings (every interval excludes zero), by \PooledAllAUC{} AUC points \PooledAllAUCCI{} and \PooledAllAcc{} accuracy points \PooledAllAccCI{} pooled.
At 192 calls, their mean accuracy exceeds GRPO's 400-call accuracy in every setting.

Witnesses reach GRPO's 400-call accuracy with at least \GPUFoldMin~times less training GPU time (Tables~\ref{tab:he} and~\ref{tab:hanoi}, and Tables~\ref{tab:gpu} and~\ref{tab:algo} in Appendices~\ref{app:statistics} and~\ref{app:effects}).

\subsection{Self-Bootstrapped Banks under a Total Budget}
\label{sec:selfbank}
\begin{figure}[t]
\centering
\includegraphics[width=\columnwidth]{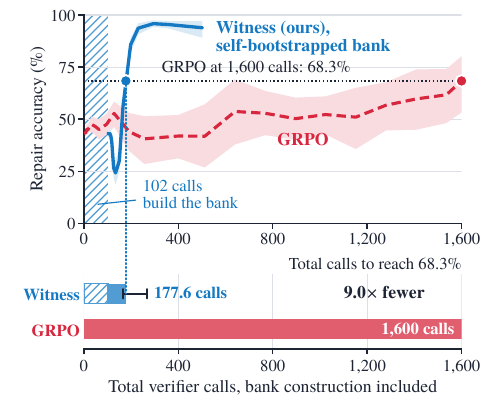}
\caption{With every bank-construction call charged, self-bootstrapped repair witnesses reach GRPO's 1,600-call accuracy after \SelfMatchTotalCalls~calls on Hanoi-Hard with \mbox{Qwen3-14B}. Top: held-out accuracy against total verifier calls, construction hatched. Bottom: total calls to reach and keep GRPO's 1,600-call mean accuracy over 12 paired seeds. Bands show 95\% seed-bootstrap intervals and the whisker a 95\% paired interval.}
\label{fig:scale}
\end{figure}

\begin{figure}[t]
\centering
\includegraphics[width=\columnwidth]{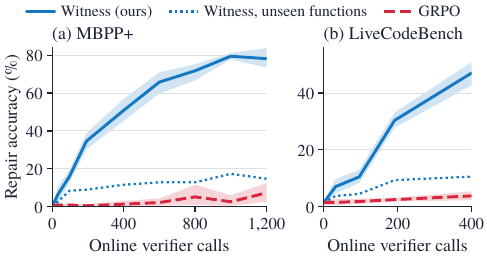}
\caption{Online witnesses learn from a cold start where GRPO stalls, with \mbox{Qwen3-1.7B} on \mbox{MBPP+} and \mbox{Qwen3-8B} on LiveCodeBench. Solid blue and dashed red show held-out bugs on functions that begin with a self-generated program, and dotted blue shows unseen functions. Bands are 95\% seed-bootstrap intervals.}
\label{fig:cold}
\end{figure}

\begin{table}[!t]
\caption{On functions whose bank begins with a self-generated program, self-bootstrapped banks outperform the curated bank on \mbox{HumanEval+}, and online witnesses outperform GRPO from a cold start. Rows score held-out bugs of these training functions. Self bank is online witnesses on the self-bootstrapped bank, and Baseline is online witnesses on the curated bank or GRPO. $\Delta$ is Self bank minus Baseline at $T=1$ with 95\% paired seed-bootstrap intervals over 12 seeds (six at 1,200 calls).}
\label{tab:selfcold}
\centering\small\setlength{\tabcolsep}{3pt}
\begin{tabular*}{\linewidth}{@{\extracolsep{\fill}}lrrrrr@{\extracolsep{0pt}\hspace{3pt}}>{\scriptsize}l@{}}
\toprule
 & \multicolumn{2}{c}{Self bank (\%)} & \multicolumn{2}{c}{Baseline (\%)} & \multicolumn{2}{c}{$\Delta$ accuracy} \\
\cmidrule(lr){2-3}\cmidrule(lr){4-5}
Model & $T{=}1$ & $T{=}0.2$ & $T{=}1$ & $T{=}0.2$ & \multicolumn{2}{c}{[95\% CI]} \\
\midrule
\multicolumn{3}{@{}l}{\textit{\mbox{HumanEval+}}} & \multicolumn{2}{c}{\textit{Curated}} & \multicolumn{2}{c}{} \\
\mbox{Qwen3-1.7B} & 70.3 & 80.5 & 63.3 & 75.9 & +7.0 & \mbox{[0.8, 13.3]} \\
\mbox{Qwen3-14B} & 86.6 & 95.5 & 82.1 & 87.8 & +4.5 & \mbox{[0.0, 8.8]} \\
Pooled &  &  &  &  & +5.8 & \mbox{[1.9, 9.6]} \\
\addlinespace[3pt]
\multicolumn{3}{@{}l}{\textit{\mbox{MBPP+}}} & \multicolumn{2}{c}{\textit{GRPO}} & \multicolumn{2}{c}{} \\
\mbox{Qwen3-1.7B} & 52.1 & 65.4 & 0.9 & 0.9 & +51.2 & \mbox{[47.9, 55.1]} \\
\mbox{SmolLM3-3B} & 56.9 & 69.4 & 26.8 & 27.1 & +30.1 & \mbox{[21.7, 38.3]} \\
\addlinespace[3pt]
\multicolumn{3}{@{}l}{\textit{\mbox{MBPP+}, 1,200 calls}} & \multicolumn{2}{c}{\textit{GRPO}} & \multicolumn{2}{c}{} \\
\mbox{Qwen3-1.7B} & 78.2 & 85.9 & 7.3 & 7.3 & +70.9 & \mbox{[65.0, 76.1]} \\
\addlinespace[3pt]
\multicolumn{3}{@{}l}{\textit{LiveCodeBench}} & \multicolumn{2}{c}{\textit{GRPO}} & \multicolumn{2}{c}{} \\
\mbox{Qwen3-8B} & 47.1 & 53.1 & 3.8 & 2.1 & +43.3 & \mbox{[39.2, 46.9]} \\
\bottomrule
\end{tabular*}

\end{table}

Repair witnesses need no correct program from outside the model.
On Hanoi-Hard, the self-bootstrapped bank of \mbox{Qwen3-14B} holds none of the curated implementations and costs 102 verifier calls to build.
With these calls charged, repair witnesses reach 93.6\% after 230 total calls (Figure~\ref{fig:scale} and Appendix~\ref{app:selfbank}), whereas GRPO reaches 68.3\% in 1,600 calls, an accuracy witnesses match with 9.0~times fewer calls and \GPUSelfFold~times less training GPU time (Table~\ref{tab:gpu}).
The self-bootstrapped bank also teaches faster than the curated one, gaining 10.9 AUC points (Appendix~\ref{app:selfbank}).
Its programs are more likely under the policy: on fixed failed proposals, base \mbox{Qwen3-14B} gives the self-bootstrapped witness a per-token negative log-likelihood (NLL) of 0.26~nats, against 0.49 for the curated one.

\begin{figure*}[t]
\centering
\includegraphics[width=\textwidth]{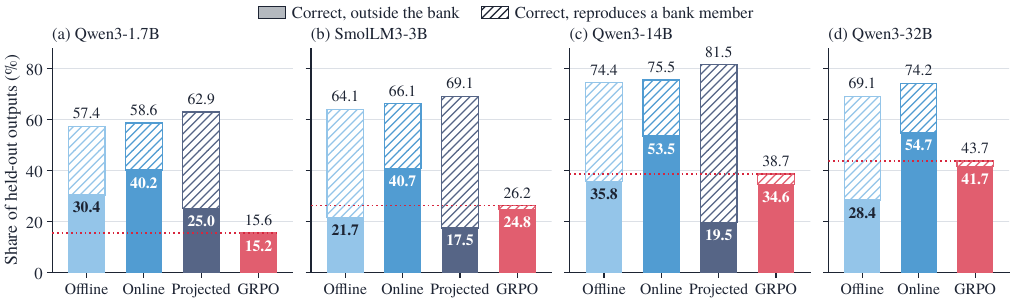}
\caption{Online witnesses exceed GRPO's total accuracy on every \mbox{HumanEval+} model with outside-bank correctness alone. Each bar splits a method's final accuracy, in its color, into passes outside the bank (solid) and reproduced bank members (hatched), and the dotted line marks GRPO's total accuracy. Offline, Online, and Projected (success-projected, run on the three smaller models) are the three witness forms.}
\label{fig:bank}
\end{figure*}
\begin{figure*}[t]
\centering
\includegraphics[width=\textwidth]{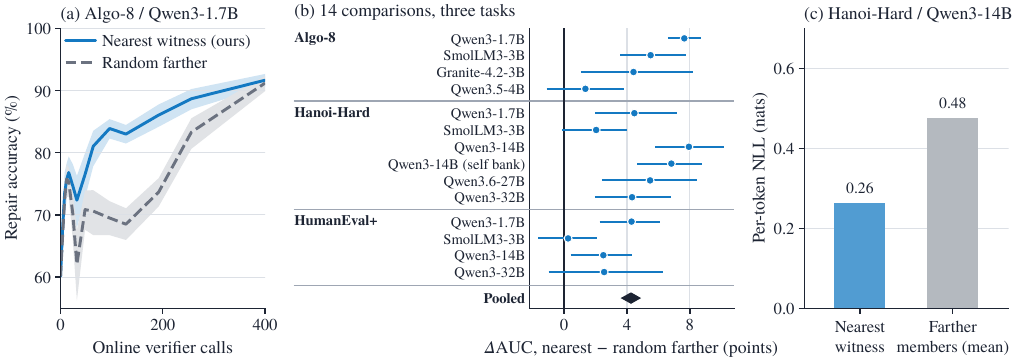}
\caption{The policy learns faster from the nearest witness, which is more likely under the policy than farther bank members. (a) Nearest against random farther witnesses on Algo-8 with \mbox{Qwen3-1.7B}. (b) Nearest minus random farther AUC in all 14 comparisons and the pooled effect (self bank marks the self-bootstrapped bank). (c) Per-token NLL of the nearest witness and the mean over farther self-bootstrapped members on fixed failed Hanoi-Hard proposals under base \mbox{Qwen3-14B}. Bands in (a) and whiskers in (b) are 95\% seed-bootstrap intervals, paired in (b).}
\label{fig:mechanism}
\end{figure*}
Natural bugs need no reference either.
On \mbox{HumanEval+}, self-bootstrapped banks for \mbox{Qwen3-1.7B} and \mbox{Qwen3-14B} more than double by admitting passing proposals and let online witnesses match GRPO's final accuracy with 12.0 and 5.0~times fewer calls (Table~\ref{tab:selfhe} in Appendix~\ref{app:selfbank}).
Without any reference, these witnesses also finish \SelfLuffyGapSmall{} and \SelfLuffyGapLarge~points above LUFFY-style guidance.
On functions that begin with a self-generated program, these banks outperform the curated bank of Table~\ref{tab:he} by 5.8~points pooled (Table~\ref{tab:selfcold}).

The same recipe works from a cold start.
With \mbox{Qwen3-1.7B} on \mbox{MBPP+} and \mbox{Qwen3-8B} on LiveCodeBench, the base policies repair \ColdBaseMbpp\% and \ColdBaseLcb\% of held-out bugs on the training functions, and online witnesses raise these rates to \ColdAllOnlineMbpp\% and \ColdAllOnlineLcb\% in 400 calls, against \ColdAllGRPOMbpp\% and \ColdAllGRPOLcb\% for GRPO (Table~\ref{tab:cold} in Appendix~\ref{app:selfbank}).
On the subset that begins with a self-generated program, they repair 52.1\% and 47.1\%, against 0.9\% and 3.8\% for GRPO (Figure~\ref{fig:cold}, Table~\ref{tab:selfcold}).
On \mbox{MBPP+}, the lead holds with a second model family and at 1,200 calls, where witnesses match GRPO's accuracy with \GPUMbppFold~times less training GPU time (Tables~\ref{tab:selfcold} and~\ref{tab:gpu}), while DAPO and Dr.~GRPO stall at 400 calls (Table~\ref{tab:cold}).

\subsection{One Family, Three Operating Points}
\label{sec:family}
Without online calls, offline witnesses finish \OffLeadMin{} to \OffLeadMax~points above every \mbox{HumanEval+} outcome-reward and reference-mixing baseline (Table~\ref{tab:he} and Table~\ref{tab:variants} in Appendix~\ref{app:training}).
Online witnesses beat offline witnesses in all ten settings of Table~\ref{tab:onoff} in Appendix~\ref{app:copy}, by 3.8~points pooled, with 1.3 to 1.9~times as much outside-bank correctness on \mbox{HumanEval+} (Figure~\ref{fig:bank}).
Unlike offline witnesses, they follow the policy's current proposals and keep its successes.

On \mbox{HumanEval+}, success-projected witnesses learn faster than online witnesses with \mbox{Qwen3-1.7B}, \mbox{SmolLM3-3B}, and \mbox{Qwen3-14B} and finish \ProjPooledOnline{} and \ProjPooledOffline~points above online and offline witnesses pooled (Figure~\ref{fig:bank} and Appendix~\ref{app:copy}).
Keeping the policy's own successes drives outside-bank correctness: online witnesses reach 40.2\% to 53.5\%, versus 17.5\% to 25.0\% for success-projected witnesses (Figure~\ref{fig:bank}).

Offline and online witnesses also compose: started from the offline-witness policy, online witnesses repair \SftLowSmall\% and \SftLowLarge\% of held-out \mbox{HumanEval+} bugs at $T=0.2$ with \mbox{Qwen3-1.7B} and \mbox{Qwen3-14B}, \SftStackSmall{} and \SftStackLarge~points above those started from the base model with the same 400 calls (Table~\ref{tab:sftrl}).

\subsection{Why the Nearest Fix and What Transfers}
\label{sec:mechanism}\label{sec:bank}
\begin{table*}[!t]
\caption{Repair skill transfers to functions absent from training and the bank. $\Delta$ is witness minus GRPO in points with 95\% paired seed-bootstrap intervals, and Base evaluates the untrained model with each row's sampling seeds.}
\label{tab:transfer}
\centering\small
\begin{tabular*}{\linewidth}{@{\extracolsep{\fill}}lllrrrrr@{\extracolsep{0pt}\hspace{3pt}}>{\scriptsize}l@{}}
\toprule
Model & Training & Unseen task & Functions & Base (\%) & Witness (\%) & GRPO (\%) & \multicolumn{2}{c}{$\Delta$ accuracy [95\% CI]} \\
\midrule
\mbox{Qwen3-1.7B} & \mbox{HumanEval+} & \mbox{HumanEvalFix} & 69 & 5.1 & 42.3 & 9.3 & +33.0 & \mbox{[24.8, 39.1]} \\
\mbox{Qwen3-1.7B} & \mbox{HumanEval+} & QuixBugs & 10 & 5.0 & 25.9 & 4.9 & +21.0 & \mbox{[18.5, 23.5]} \\
\mbox{SmolLM3-3B} & \mbox{HumanEval+} & QuixBugs & 10 & 19.6 & 34.9 & 18.8 & +16.1 & \mbox{[12.9, 19.2]} \\
\mbox{Qwen3-1.7B} & Hanoi-Hard & \mbox{HumanEvalFix} & 69 & 5.0 & 34.6 & 15.7 & +18.9 & \mbox{[11.1, 25.2]} \\
\mbox{Qwen3-1.7B} & Hanoi-Hard & QuixBugs & 10 & 5.5 & 22.4 & 12.6 & +9.8 & \mbox{[4.8, 15.0]} \\
\mbox{Qwen3-14B} & Hanoi-Hard & QuixBugs & 10 & 31.7 & 66.2 & 43.5 & +22.7 & \mbox{[12.6, 33.8]} \\
\bottomrule
\end{tabular*}

\end{table*}
Witness gains come from the supervised term and its choice of target.
Without that term, REINFORCE finishes \ReinforceGapSmall{} and \ReinforceGapLarge~points below repair witnesses on Algo-8 with \mbox{Qwen3-1.7B} and \mbox{Qwen3.5-4B} (Table~\ref{tab:algo} in Appendix~\ref{app:effects}), and on \mbox{HumanEval+} it only matches GRPO, \ReinforceHeGapSmall{} and \ReinforceHeGapLarge~points below online witnesses with \mbox{Qwen3-1.7B} and \mbox{Qwen3-14B} (Table~\ref{tab:ceonly}).
That term alone keeps \CEKeepAucSmall\% and \CEKeepAucLarge\% of the AUC lead of online witnesses over GRPO and \CEKeepAccSmall\% and \CEKeepAccLarge\% of their $T=1$ lead, and the policy-gradient term adds \CEPgAuc~AUC points \CEPgAucCI{} on \mbox{Qwen3-14B}.

Nearest witnesses beat random farther ones by 7.6 AUC points on Algo-8 with \mbox{Qwen3-1.7B} (Figure~\ref{fig:mechanism}a) and lead in all 14 comparisons across seven models and three tasks, including the self-bootstrapped bank, by 4.3~points pooled (Figure~\ref{fig:mechanism}b and Table~\ref{tab:choice} in Appendix~\ref{app:effects}).
Proximity in tokens is proximity to the policy.
On fixed failed Hanoi-Hard proposals, base \mbox{Qwen3-14B} gives the nearest self-bootstrapped witness a per-token NLL of 0.26~nats, against 0.48 for farther members (Figure~\ref{fig:mechanism}c).
Likelihood witnesses, which retrieve the bank member that the current policy finds most likely, learn faster still, by \LikAucSmall{} and \LikAucLarge~AUC points on \mbox{HumanEval+} with \mbox{Qwen3-1.7B} and \mbox{Qwen3-14B}, and finish level at $T=0.2$ (Table~\ref{tab:ceonly}).
Edit distance picks the same target for most failures without running the policy on the bank.

Beyond the training functions, witness-trained policies outperform GRPO in all six transfer evaluations on HumanEvalFix and QuixBugs (Table~\ref{tab:transfer}), and on QuixBugs, \mbox{Qwen3-14B} trained on Hanoi-Hard reaches 66.2\% with online witnesses, against 57.5\% with offline witnesses.

\section{Related Work}
\label{sec:related}
Policy-gradient and preference methods learn from sequence-level outcomes \citep{reinforce,ppo,dpo}, and CodeRL, GRPO, and DeepSeek-R1 bring them to code and reasoning \citep{coderl,deepseekmath,deepseekr1}.
DAPO resamples uniform-reward groups and Dr.~GRPO removes an optimization bias of GRPO \citep{dapo,drgrpo}. RLTF and StepCoder add unit-test and compiler feedback \citep{rltf,stepcoder}, and SWE-RL rewards similarity to the oracle patch \citep{swerl}.
Repair witnesses keep the binary verifier and every sampled group but give failures a program target, not a scalar reward.

Self-Debug, Self-Refine, and Reflexion rely on explanations or reflective feedback \citep{selfdebug,selfrefine,reflexion}, RLEF and SCoRe train models to use execution feedback and to self-correct across turns \citep{rlef,score}, and SDPO distills a feedback-conditioned self-teacher into the policy \citep{sdpo}.
Repair witnesses use binary feedback and retrieve verified targets without extra calls or turns.

DAgger gathers supervision at states the learner visits, hindsight experience replay learns from unsuccessful episodes \citep{dagger,her}, and optimal completion distillation derives targets from edit distance to a reference \citep{ocd}.
STaR, ReST, \mbox{ReST-EM}, RAFT, and self-imitation learning train on the model's own selected successes \citep{star,rest,restem,raft,sil}, and LUFFY (Section~\ref{sec:he}), ReLIFT, CHORD, HPT, and BREAD mix expert traces into on-policy learning \citep{luffy,relift,chord,hpt,bread}.
Repair witnesses retrieve failure targets from the bank and keep the policy's successes, so supervision tracks the learner without expert demonstrations.

Concurrent with our work, projected distillation has a stronger model minimally edit each failed attempt into a correct answer, trains on the edited spans with a contrastive loss and on the policy's own correct answers, and shows that targets farther from the policy cost more of its prior abilities \citep{projdistill}.
Its analysis of edit size and ours of witness distance arrive independently at the same principle: a failure should learn from a close, correct target.
Repair witnesses retrieve that target from verified programs, with no teacher queried, no rewrite to verify, and all cost counted in verifier calls.

Clara and Refactory repair student programs by aligning them with the closest correct submission in a pool, a precedent for the nearest-fix principle \citep{clara,refactory}. Conversational, completion-guided, adapter-based, and repository-level repair are complementary \citep{chatrepair,repilot,repairllama,swebench,agentless}.

\section{Conclusion}
\label{sec:conclusion}
Repair witnesses turn failed verifier calls into supervision by matching each failure to the nearest program in a bank of verified implementations, and a bank built from the model's own passing proposals needs no teacher or reference solution (Section~\ref{sec:selfbank}).
Across offline, online, and success-projected forms, witnesses outpace every outcome-reward and reference-mixing baseline within the same call budgets.
The policy learns faster from the nearest verified program than from a farther one, and what it learns extends beyond the bank and the training functions.
A verifier can do more than judge an attempt: with the programs it has already accepted, every failure it reports also points to the nearest fix.

\label{main-end}
\section*{Impact Statement}
This work improves the verifier efficiency of reinforcement learning for code repair.
The experiments use public programming benchmarks and openly released models, with no new collection of user data.
More efficient repair training can make automated debugging and software maintenance more accessible.
Generated repairs should undergo the testing and review appropriate to their deployment context.

\bibliography{refs}
\bibliographystyle{icml2026}
\clearpage
\onecolumn
\appendix
\section{Data and Evaluation Sets}
\label{app:data}
Table~\ref{tab:data} summarizes every task's data and evaluation sets, shared by all methods for a model.

\begin{table}[!htb]
\caption{Data and evaluation sets. Training inputs are equally weighted. Distinct counts different defective programs, Bank counts canonically distinct verified implementations, and Model's own counts those sampled from the model being trained. Hanoi-Hard and Algo-8 share one dataset across their models, and on \mbox{MBPP+} and LiveCodeBench, Held-out has one held-out bug for every training and unseen function. Generations/point is held-out inputs times samples per input at each evaluation point.}
\centering\small
\begin{tabular*}{\linewidth}{@{\extracolsep{\fill}}llrrrrrr@{}}
\toprule
Task & Model & Training inputs & Distinct & Held-out & Bank & Model's own & Generations/point \\
\midrule
\mbox{HumanEval+} & \mbox{Qwen3-1.7B} & 160 & 142 & 40 & 123 & 84 (68.3\%) & $40\times2=80$ \\
 & \mbox{SmolLM3-3B} & 160 & 156 & 40 & 121 & 82 (67.8\%) & $40\times2=80$ \\
 & \mbox{Qwen3-14B} & 108 & 103 & 27 & 80 & 54 (67.5\%) & $27\times3=81$ \\
 & \mbox{Qwen3-32B} & 108 & 103 & 27 & 102 & 75 (73.5\%) & $27\times3=81$ \\
\addlinespace[3pt]
Hanoi-Hard & Eight models & 40 & 40 & 40 & 12 & -- & $40\times2=80$ \\
\addlinespace[3pt]
Algo-8 & Four models & 160 & 160 & 160 & 48 & -- & $160\times2=320$ \\
\addlinespace[3pt]
\mbox{MBPP+} & \mbox{Qwen3-1.7B} & 412 & 364 & 129 & 87 & 87 (100.0\%) & $129\times1=129$ \\
 & \mbox{SmolLM3-3B} & 612 & 546 & 191 & 248 & 248 (100.0\%) & $191\times1=191$ \\
\addlinespace[3pt]
LiveCodeBench & \mbox{Qwen3-8B} & 640 & 624 & 220 & 191 & 191 (100.0\%) & $220\times1=220$ \\
\bottomrule
\end{tabular*}

\label{tab:data}
\end{table}

\subsection{\mbox{HumanEval+} Repair}
HumanEvalFix comes from the Python split of HumanEvalPack \citep{octopack}, with one human-written defect for each HumanEval function, and EvalPlus supplies the corresponding extended \mbox{HumanEval+} tests \citep{evalplus}.
We align the two datasets by function identifier and entry point.
A candidate function needs a passing reference implementation within the execution time limit, a prompt and output that fit the length caps, a nonempty verified bank, and at least two distinct natural training defects.

For each model and function, 16 repair attempts on the human-written defect estimate base accuracy, which must be at most 75\% for eligibility.
Selection keeps up to 40 eligible functions with base accuracy in $(0\%, 50\%]$, those closest to 25\% first, and fills the remaining places one at a time from the other eligible functions, each time taking the function that brings the mean base accuracy of the set closest to 25\%.
This gives 40 functions for \mbox{Qwen3-1.7B} and \mbox{SmolLM3-3B} and all 27 eligible functions for \mbox{Qwen3-14B} and \mbox{Qwen3-32B}.

A natural defect parses, defines the entry point, ends without truncation, and fails the functional verifier.
Distinctness uses canonical token sequences. Each function's training inputs cycle over its training defects, which are canonically distinct from the bank, the held-out defect, and one another.
With \mbox{Qwen3-14B} and \mbox{Qwen3-32B}, natural failures of smaller Qwen3 models on the same specifications fill out the training defects where the model's own failures fall short, and the bank still holds only the model's own programs and the reference.

HumanEvalPack is distributed under the MIT license and \mbox{HumanEval+} under Apache-2.0.
Verification runs each program in a resource-limited subprocess.
All methods for a task share the same prompt, decoding, and verifier.

\subsection{\mbox{MBPP+} and LiveCodeBench}
\mbox{MBPP+} uses the EvalPlus tests for MBPP \citep{mbpp,evalplus}, and LiveCodeBench uses the functional problems of release v6 \citep{livecodebench}.
The model being trained writes 16 programs per function from the specification.
Failures that satisfy the natural-defect conditions above become defects, and passes become bank members, at most six per function after canonical deduplication.
Eligibility requires at least three distinct defects: one held out for evaluation and up to four for training. As on \mbox{HumanEval+}, each function's four training inputs cycle over its training defects.
A fixed hash assigns \ColdUnseenShare\% of the eligible functions to the unseen split (\ColdUnseenMbpp{} of \ColdEligibleMbpp{} and \SmolColdUnseen{} of \ColdEligibleSmol{} on \mbox{MBPP+} with \mbox{Qwen3-1.7B} and \mbox{SmolLM3-3B}, \ColdUnseenLcb{} of \ColdEligibleLcb{} on LiveCodeBench), whose defects and programs never enter training or the bank. Training keeps up to \ColdTrainCap{} of the remaining functions, those with base repair accuracy closest to 25\% first.
\mbox{MBPP+} is distributed under Apache-2.0 and the LiveCodeBench code under the MIT license.

\subsection{Hanoi-Hard and Algo-8}
Hanoi-Hard combines \texttt{cyclic\_hanoi} and \texttt{linear\_hanoi}.
Each function contributes 20 training defects, 20 held-out defects, and six canonically distinct verified implementations to the shared curated bank.

Algo-8 uses \texttt{bitcount}, \texttt{gcd}, \texttt{get\_factors}, \texttt{flatten}, \texttt{is\_valid\_parenthesization}, \texttt{pascal}, \texttt{hanoi}, and \texttt{to\_base}.
Single-node mutations of verified implementations that fail the tests are deduplicated canonically and split at random into 20 training and 20 held-out defects per function.
QuixBugs supplies the public algorithmic functions under the MIT license \citep{quixbugs}.

\subsection{Unseen-Function Transfer}
\label{app:transfer}
The 69 HumanEvalFix functions, the same for both HumanEvalFix rows of Table~\ref{tab:transfer}, lie outside the \mbox{Qwen3-1.7B} \mbox{HumanEval+} training set and banks and have base accuracy at most 75\%.
The 10 QuixBugs functions exclude the Hanoi family and every function whose entry point or reference implementation appears in training or in a bank.
Each function receives 32~generations per final policy, and the base model uses the same input, decoding, and verifier.

\subsection{Verifier Budget and Bank Construction}
\label{app:budget}
The online budget is 400 calls for every online method except the 1,200-call runs on \mbox{MBPP+} and the 1,600-call GRPO run on Hanoi-Hard.
Every curated bank and the self-bootstrapped Hanoi-Hard bank are built once before training and shared across all witness runs that use them.
On \mbox{HumanEval+}, \mbox{MBPP+}, and LiveCodeBench, the initial bank and the natural training defects come from the same specification-conditioned samples, so the bank adds no verifier calls beyond building the shared training set.
The self-bootstrapped Hanoi-Hard bank takes the proposals that base \mbox{Qwen3-14B} samples for the training defects before its first update: 96 proposals, each verified once, of which 23 pass.
After canonical deduplication and removal of every program that coincides with a curated member, three implementations per function remain, and rechecking these six brings construction to 102 verifier calls.
Offline witnesses take their supervision from the inputs whose function has a bank member.

\section{Canonicalization and Witness Selection}
\label{app:canonical}
Canonicalization makes target distance insensitive to formatting and surface variation.
It parses the program, removes comments and docstrings, normalizes supported type annotations and imports, and deterministically renames local bindings.
Tokens follow the normalized syntax tree, and indentation and dedentation keep their token kinds without whitespace strings.
The canonicalizer for \mbox{HumanEval+}, \mbox{MBPP+}, and LiveCodeBench keeps top-level entry names.
The Hanoi-Hard and Algo-8 canonicalizer supports functions without lambdas, classes, global or nonlocal declarations, or nested-scope shadowing.

For a proposal with a supported canonical form, the selector computes exact Levenshtein distance to every bank member and samples uniformly among the members at minimum distance.
A proposal that cannot be canonicalized falls back to the first bank member under every witness rule: the reference implementation of a curated bank or the earliest verified program of a self-bootstrapped bank.
Online witnesses keep a passing proposal's sampled token sequence as its target (Algorithm~\ref{alg:update}).

\begin{algorithm}[!htb]
\caption{One update with online repair witnesses}
\label{alg:update}
\begin{algorithmic}[1]
\REQUIRE policy $\policy$, training inputs $\mathcal{D}$, verified banks $\{\bank_f\}$, inputs per update $m$, group size $G$, weight $\lambda$
\STATE Draw inputs $x_1,\dots,x_m$ at random from $\mathcal{D}$
\FOR{$j=1,\dots,m$}
\STATE Sample proposals $y_{j1},\dots,y_{jG}\sim\policy(\cdot\mid x_j)$
\STATE Verify $r_{jk}\gets V_{f_j}(y_{jk})$ for each $k$ \COMMENT{$G$ calls}
\STATE $\widehat A_{jk}\gets r_{jk}-\frac{1}{G}\sum_{k'}r_{jk'}$
\FOR{$k=1,\dots,G$}
\IF{$r_{jk}=1$}
\STATE $t_{jk}\gets y_{jk}$ \COMMENT{own success}
\ELSIF{$\bank_{f_j}\neq\varnothing$}
\STATE $t_{jk}\gets w^\star_{f_j}(y_{jk})$ \COMMENT{witness, no call}
\ELSE
\STATE $t_{jk}\gets\varnothing$ \COMMENT{no supervised term}
\ENDIF
\ENDFOR
\ENDFOR
\STATE $\mathcal{L}\gets\frac{1}{mG}\sum_{j,k}\bigl[-\widehat A_{jk}\,\bar\ell_\theta(y_{jk}\mid x_j)+\lambda\CE_\theta(t_{jk}\mid x_j)\bigr]$
\STATE Clip $\nabla_\theta\mathcal{L}$ to norm one and take an AdamW step
\STATE With a growing bank, add every passing $y_{jk}$ that is canonically new to $\bank_{f_j}$ \COMMENT{no call}
\end{algorithmic}
\end{algorithm}

For the random farther witness, each seed draws one far anchor per function before training from a pool fixed with the data: on Algo-8, the members strictly farther than the nearest member from every defect of the function; on the curated Hanoi-Hard bank, one designated implementation; and on \mbox{HumanEval+} and the self-bootstrapped Hanoi-Hard bank, the member with the largest total canonical token edit distance to the function's training defects.
When the anchor is strictly farther than the nearest member for the current failure, the control samples uniformly from all strictly farther members, and otherwise it uses the nearest member.

\section{Training Configuration and Baselines}
\label{app:training}
We use low-rank adaptation (LoRA) with zero dropout on every linear layer except the output head (on \mbox{Muse Glimmer-30B}, the attention and MLP projections of its language model) and keep the base weights fixed in bfloat16 \citep{lora}.
Table~\ref{tab:hyper} gives the common configuration.
Experiments run on NVIDIA H100, RTX~PRO~6000, and L4 GPUs.
The Qwen models follow their released instruction interfaces with thinking disabled \citep{qwen3,qwen35,qwen36}, and generation uses the same chat-template setting within each comparison.
Each input is one user message: an instruction, then \texttt{Specification:} with the function specification and \texttt{Buggy code:} with the defective implementation.
On \mbox{HumanEval+} and \mbox{MBPP+}, the instruction reads ``Repair the following Python function. Preserve its name and signature. Return the complete corrected Python code, including any imports and helper functions, without prose, tests, or Markdown.''
LiveCodeBench asks for a repaired Python solution that preserves \texttt{class Solution} and its method signature, and Hanoi-Hard and Algo-8 ask for function definitions without imports and list the available builtins.
We use AdamW without weight decay, clip the gradient norm at one, and take one optimizer step per batch.
The witness objective uses centered rewards without standard-deviation normalization or a KL term.

\begin{table}[!htb]
\caption{Common training configuration. GRPO samples one input with four proposals per update, and offline witnesses take four supervised targets per update without online sampling. \mbox{MBPP+} raises the prompt and completion token caps to \CapPromptMbpp{} and \CapCompletionMbpp{}, and LiveCodeBench to \CapPromptLcb{} and \CapCompletionLcb{}.}
\centering\small
\begin{tabular*}{\linewidth}{@{\extracolsep{\fill}}lrlr@{}}
\toprule
Parameter & Value & Parameter & Value \\
\midrule
LoRA rank & 16 & Supervised weight $\lambda$ & 1.0 \\
LoRA $\alpha$ & 32 & Updates & 100 \\
Learning rate & 0.0001 & Batch inputs (witness/RFT) & 2 \\
Prompt token cap & 768 & Proposals per input (witness/RFT) & 2 \\
Completion token cap & 384 & GRPO KL weight & 0.04 \\
Sampling temperature & 1.0 & Bootstrap replicates & 10,000 \\
\bottomrule
\end{tabular*}

\label{tab:hyper}
\end{table}

For GRPO, the advantage is the centered binary reward divided by the group's population standard deviation plus a small stabilizer.
The KL term uses the adapter-disabled base policy as reference and the nonnegative sampled estimator
\[
 k_3=\rho_t-\log\rho_t-1,\qquad
 \rho_t=\frac{\pi_{\mathrm{ref}}(z_t\mid x,z_{<t})}{\policy(z_t\mid x,z_{<t})},
\]
for each continuation token $z_t$.
Its mean over continuation tokens times the KL weight is added to the policy-gradient loss.
On \mbox{MBPP+} and \mbox{HumanEval+}, DAPO-style dynamic sampling draws one group of four per step and skips the update whenever all four rewards agree, still charging the group's calls.
Dr.~GRPO drops the standard-deviation normalization and divides each sequence log-likelihood by a fixed length instead of the response length.
RFT uses the same input sampling as repair witnesses and applies cross-entropy only to successful programs, with failed samples kept in the batch denominator.

\begin{table}[!htb]
\caption{Online witnesses outperform every GRPO variant on \mbox{HumanEval+}, and no tested learning rate or batch shape beats standard GRPO. Top: larger-group GRPO ($G=8$), KL-free GRPO (no KL), DAPO, Dr.~GRPO, and random-bank SFT (no online calls) below each model's GRPO row of Table~\ref{tab:he}, with metrics following Table~\ref{tab:he} at $T=1$ over 12 seeds; DAPO and Dr.~GRPO are evaluated after 400 calls. Zero adv.\ is the percentage of updates in which every proposal of the group has the same reward (for DAPO, which skips such groups, of sampled groups). Bottom: GRPO with learning rate (LR) $3\times10^{-5}$ or $3\times10^{-4}$, or with two inputs of four proposals ($2\times4$) per update for 50 updates, against standard GRPO (LR $10^{-4}$, one input of four proposals, 100 updates) at $T=1$ with 95\% paired seed-bootstrap intervals over six of the 12 seeds, so the Standard rows differ from the GRPO rows above.}
\centering\small\setlength{\tabcolsep}{4pt}
\begin{tabular*}{\linewidth}{@{\extracolsep{\fill}}llrrr@{\extracolsep{0pt}\hspace{3pt}}>{\scriptsize}l@{\extracolsep{\fill}\hspace{2\tabcolsep}}r@{\extracolsep{0pt}\hspace{3pt}}>{\scriptsize}l@{\extracolsep{\fill}\hspace{2\tabcolsep}}r@{\extracolsep{0pt}\hspace{3pt}}>{\scriptsize}l@{\extracolsep{\fill}\hspace{2\tabcolsep}}r@{\extracolsep{0pt}\hspace{3pt}}>{\scriptsize}l@{}}
\toprule
 &  & Accuracy & Zero adv. & \multicolumn{2}{c}{$\Delta$AUC} & \multicolumn{2}{c}{$\Delta$ accuracy} & \multicolumn{2}{c}{Fewer} & \multicolumn{2}{c}{Less GPU} \\
Model & Method & (\%) & (\%) & \multicolumn{2}{c}{[95\% CI]} & \multicolumn{2}{c}{[95\% CI]} & \multicolumn{2}{c}{calls ($\times$)} & \multicolumn{2}{c}{time ($\times$)} \\
\midrule
\mbox{Qwen3-1.7B} & GRPO & 15.6 & 98.2 & +17.6 & \mbox{[14.8, 20.4]} & +43.0 & \mbox{[32.7, 51.0]} & 3.3 & \mbox{[2.7, 3.7]} & 1.8 & \mbox{[1.6, 2.1]} \\
 & GRPO ($G=8$) & 14.5 & 99.2 & +18.5 & \mbox{[16.7, 20.2]} & +44.1 & \mbox{[39.3, 48.6]} & 3.4 & \mbox{[3.0, 3.7]} & 1.3 & \mbox{[1.2, 1.4]} \\
 & GRPO (no KL) & 13.9 & 93.8 & +16.0 & \mbox{[10.0, 21.7]} & +44.7 & \mbox{[34.3, 53.2]} & 3.4 & \mbox{[2.8, 5.7]} & 2.4 & \mbox{[1.9, 4.3]} \\
 & DAPO & 17.5 & 99.1 & -- &  & +41.1 & \mbox{[32.8, 48.1]} & 3.1 & \mbox{[2.7, 3.6]} & -- &  \\
 & Dr.~GRPO & 15.5 & 99.2 & -- &  & +43.1 & \mbox{[35.4, 49.1]} & 3.3 & \mbox{[2.9, 3.7]} & -- &  \\
 & Random-bank SFT & 50.4 & -- & +2.1 & \mbox{[0.4, 3.8]} & +8.2 & \mbox{[6.0, 10.4]} & -- &  & -- &  \\
\addlinespace[3pt]
\mbox{SmolLM3-3B} & GRPO & 26.2 & 78.0 & +15.4 & \mbox{[12.4, 18.3]} & +39.9 & \mbox{[31.4, 48.3]} & 3.4 & \mbox{[3.0, 4.4]} & 3.2 & \mbox{[2.8, 4.0]} \\
 & GRPO ($G=8$) & 30.6 & 63.5 & +14.2 & \mbox{[11.1, 17.0]} & +35.5 & \mbox{[27.8, 42.4]} & 3.0 & \mbox{[2.5, 3.5]} & 2.0 & \mbox{[1.7, 2.3]} \\
 & GRPO (no KL) & 18.4 & 80.4 & +22.2 & \mbox{[18.5, 26.0]} & +47.7 & \mbox{[38.5, 56.8]} & 4.6 & \mbox{[3.3, 6.4]} & 4.9 & \mbox{[3.7, 6.3]} \\
 & Random-bank SFT & 59.7 & -- & +2.5 & \mbox{[0.9, 4.2]} & +6.4 & \mbox{[0.9, 12.0]} & -- &  & -- &  \\
\addlinespace[3pt]
\mbox{Qwen3-14B} & GRPO & 38.7 & 84.8 & +23.2 & \mbox{[18.7, 28.4]} & +36.8 & \mbox{[28.9, 45.6]} & 3.4 & \mbox{[2.7, 4.3]} & 2.3 & \mbox{[1.8, 3.2]} \\
 & GRPO ($G=8$) & 40.4 & 83.5 & +27.3 & \mbox{[24.0, 30.5]} & +35.1 & \mbox{[29.1, 40.9]} & 3.3 & \mbox{[2.8, 3.8]} & 1.4 & \mbox{[1.3, 1.6]} \\
 & GRPO (no KL) & 42.9 & 83.7 & +17.8 & \mbox{[14.3, 22.2]} & +32.6 & \mbox{[24.1, 43.1]} & 2.9 & \mbox{[2.4, 4.0]} & 2.2 & \mbox{[1.7, 3.2]} \\
 & Random-bank SFT & 72.0 & -- & +2.3 & \mbox{[0.0, 4.7]} & +3.5 & \mbox{[$-$3.1, 10.0]} & -- &  & -- &  \\
\bottomrule
\end{tabular*}

\par\vspace{8pt}
\begin{tabular*}{\linewidth}{@{\extracolsep{\fill}}llrrr@{\extracolsep{0pt}\hspace{3pt}}>{\scriptsize}l@{\extracolsep{\fill}\hspace{2\tabcolsep}}r@{\extracolsep{0pt}\hspace{3pt}}>{\scriptsize}l@{\extracolsep{\fill}\hspace{2\tabcolsep}}r@{\extracolsep{0pt}\hspace{3pt}}>{\scriptsize}l@{\extracolsep{\fill}\hspace{2\tabcolsep}}r@{\extracolsep{0pt}\hspace{3pt}}>{\scriptsize}l@{}}
\toprule
 &  &  &  & \multicolumn{4}{c}{Online witness $-$ row} & \multicolumn{4}{c}{Row $-$ standard GRPO} \\
\cmidrule(lr){5-8}\cmidrule(lr){9-12}
 &  & AUC & Accuracy & \multicolumn{2}{c}{$\Delta$AUC} & \multicolumn{2}{c}{$\Delta$ accuracy} & \multicolumn{2}{c}{$\Delta$AUC} & \multicolumn{2}{c}{$\Delta$ accuracy} \\
Model & GRPO setting & (\%) & (\%) & \multicolumn{2}{c}{[95\% CI]} & \multicolumn{2}{c}{[95\% CI]} & \multicolumn{2}{c}{[95\% CI]} & \multicolumn{2}{c}{[95\% CI]} \\
\midrule
\mbox{Qwen3-1.7B} & Standard & 15.8 & 22.1 & +17.9 & \mbox{[13.4, 22.4]} & +35.6 & \mbox{[18.5, 50.0]} & -- &  & -- &  \\
 & LR $3{\times}10^{-5}$ & 15.8 & 20.8 & +17.9 & \mbox{[10.6, 24.3]} & +36.9 & \mbox{[15.6, 55.6]} & 0.0 & \mbox{[$-$6.5, 7.4]} & $-$1.3 & \mbox{[$-$15.8, 14.4]} \\
 & LR $3{\times}10^{-4}$ & 9.7 & 4.8 & +24.0 & \mbox{[21.3, 25.8]} & +52.9 & \mbox{[47.5, 57.3]} & $-$6.1 & \mbox{[$-$10.2, $-$2.0]} & $-$17.3 & \mbox{[$-$29.0, $-$6.7]} \\
 & $2{\times}4$, 50 updates & 12.0 & 15.6 & +21.7 & \mbox{[19.0, 24.6]} & +42.1 & \mbox{[36.0, 47.3]} & $-$3.8 & \mbox{[$-$7.5, $-$0.3]} & $-$6.5 & \mbox{[$-$19.0, 5.6]} \\
\addlinespace[3pt]
\mbox{Qwen3-14B} & Standard & 25.0 & 34.8 & +28.7 & \mbox{[23.4, 35.3]} & +43.2 & \mbox{[33.7, 56.4]} & -- &  & -- &  \\
 & LR $3{\times}10^{-5}$ & 20.0 & 24.3 & +33.7 & \mbox{[30.1, 38.6]} & +53.7 & \mbox{[45.5, 62.8]} & $-$5.0 & \mbox{[$-$9.4, 1.2]} & $-$10.5 & \mbox{[$-$21.4, 2.5]} \\
 & LR $3{\times}10^{-4}$ & 19.8 & 8.6 & +33.9 & \mbox{[27.6, 41.4]} & +69.4 & \mbox{[57.6, 80.9]} & $-$5.2 & \mbox{[$-$13.5, 4.0]} & $-$26.2 & \mbox{[$-$38.5, $-$12.1]} \\
 & $2{\times}4$, 50 updates & 23.6 & 30.7 & +30.1 & \mbox{[25.6, 34.1]} & +47.3 & \mbox{[34.2, 58.8]} & $-$1.4 & \mbox{[$-$7.9, 8.2]} & $-$4.1 & \mbox{[$-$18.7, 20.2]} \\
\bottomrule
\end{tabular*}

\label{tab:variants}
\end{table}

In Table~\ref{tab:variants}, random-bank SFT trains as offline witnesses do but draws each target uniformly from the bank, and its AUC places update $k$ at $4k$ calls.

LUFFY-style guidance follows \citet{luffy} within the GRPO configuration.
Every update adds the function's EvalPlus reference implementation to the group of four proposals with a reward of one and no verifier call, normalizes advantages over the mixed group, and applies the shaping $\phi(p)=p/(p+\gamma)$ with $\gamma=\LuffyGamma$ to each token probability $p$ of the reference under the policy.
The KL weight is zero, as in LUFFY.
When all four proposals fail, the reference gives each proposal an advantage of \LuffyAdv, so the update lowers the likelihood of every program the policy wrote.
Over the four models, \LuffyZeroAll{} of the \LuffyRunsAll{} LUFFY-style runs end at zero accuracy, against \LuffyZeroGRPO{} of the \LuffyRunsAll{} GRPO runs, whereas every online witness run finishes above its starting accuracy.

\FloatBarrier
\section{Statistical Definitions}
\label{app:statistics}
For a seed's held-out accuracy curve $a(c)$ at online call count $c$, the normalized AUC is
\[
\operatorname{AUC}_{C}=\frac{1}{C}\int_0^C a(c)\,dc,
\]
computed by trapezoidal integration on the evaluation grid.
Method differences are computed within each seed before averaging or resampling.

Fewer calls compares repair witnesses with a baseline at the accuracy the baseline reaches with its full budget.
The target is the mean final accuracy of the baseline over paired seeds.
For each witness seed, the matching cost is the linearly interpolated call count after which held-out accuracy stays at or above the target, or 400 calls if it does not.
Fewer calls divides the baseline's budget of 400, 1,200, or 1,600 calls by the mean matching cost.
On the total-call axis of the self-bootstrapped bank, the 102 construction calls are added to the matching cost of every seed.

We resample paired training seeds with replacement for 10,000 bootstrap replicates and report 95\% paired seed-bootstrap intervals using percentile bounds.
A pooled effect resamples paired seeds independently within each setting and weights settings equally.
Task sets and generated programs stay fixed within each replicate, and every replicate recomputes the fewer-calls target from its resampled baseline seeds.
Analyses use 12 paired seeds with five exceptions.
The comparison of nearest and random farther witnesses on Algo-8 with \mbox{Qwen3-1.7B} has 53, the 1,200-call \mbox{MBPP+} comparison and the GRPO sweep have six each, the offline-witness start with \mbox{Qwen3-1.7B} has \SftPairsSmall{}, and training GPU time against LUFFY-style guidance with \mbox{Qwen3-1.7B} has \LuffyGPUPairs{}, the seeds whose two runs share a GPU type.

\begin{table}[!htb]
\caption{Repair witnesses reach the accuracy of GRPO with less training GPU time. Update ratio is the mean per-seed ratio of update time, witness over baseline, and GPU minutes give the witness's training time up to its matching cost and the baseline's for its full budget. $^\dagger$ marks rows whose witness runs, and any 400-call GRPO runs they pair with, kept only their total time. Fewer calls, Less GPU time, and $>$ follow Section~\ref{sec:metrics}, with 95\% paired seed-bootstrap intervals over 12 seeds (six for the 1,200-call row).}
\centering\small\setlength{\tabcolsep}{4.5pt}
\begin{tabular*}{\linewidth}{@{\extracolsep{\fill}}lllrrr@{\extracolsep{0pt}\hspace{3pt}}>{\scriptsize}l@{\extracolsep{\fill}\hspace{2\tabcolsep}}r@{\extracolsep{0pt}\hspace{3pt}}>{\scriptsize}l@{}}
\toprule
 &  &  & Update & GPU & \multicolumn{2}{c}{Fewer} & \multicolumn{2}{c}{Less GPU} \\
Task & Model & Baseline & ratio & minutes & \multicolumn{2}{c}{calls ($\times$)} & \multicolumn{2}{c}{time ($\times$)} \\
\midrule
Hanoi-Hard & \mbox{Qwen3-14B} & GRPO & 1.19\rlap{$^\dagger$} & 8.3 / 24.0 & 5.4 & \mbox{[4.7, 6.5]} & 2.9 & \mbox{[2.5, 3.5]} \\
 & \mbox{Qwen3.6-27B} & GRPO & 1.08\rlap{$^\dagger$} & 11.9 / 69.2 & 11.1 & \mbox{[2.9, 26.6]} & 5.8 & \mbox{[2.0, 12.0]} \\
 & \mbox{Qwen3-32B} & GRPO & 1.09\rlap{$^\dagger$} & 12.6 / 41.6 & 5.7 & \mbox{[3.8, 6.6]} & 3.3 & \mbox{[2.4, 3.8]} \\
\addlinespace[3pt]
 & \mbox{Qwen3-14B} & GRPO, 1,600 calls & 1.30\rlap{$^\dagger$} & 10.7 / 98.3 & 15.8 & \mbox{[6.8, 19.0]} & 9.2 & \mbox{[4.9, 10.4]} \\
 & \mbox{Qwen3-14B}, self-bootstrapped bank & GRPO, 1,600 calls & 1.16 & 15.1 / 98.3 & 9.0 & \mbox{[6.0, 9.6]} & 6.5 & \mbox{[4.7, 6.9]} \\
\addlinespace[3pt]
\mbox{MBPP+} & \mbox{Qwen3-1.7B} & GRPO & 1.40\rlap{$^\dagger$} & 1.2 / 12.3 & $>$16 &  & $>$5.8 &  \\
 &  & GRPO, 1,200 calls & 1.40\rlap{$^\dagger$} & 3.2 / 39.6 & 29.4 & \mbox{[15.6, 85.8]} & 12.2 & \mbox{[7.5, 29.4]} \\
 & \mbox{SmolLM3-3B} & GRPO & 0.70\rlap{$^\dagger$} & 4.3 / 27.3 & 7.2 & \mbox{[5.4, 9.4]} & 6.3 & \mbox{[4.9, 7.9]} \\
\addlinespace[3pt]
LiveCodeBench & \mbox{Qwen3-8B} & GRPO & 1.63\rlap{$^\dagger$} & 5.6 / 33.4 & 18.0 & \mbox{[9.1, 43.3]} & 5.9 & \mbox{[3.3, 12.0]} \\
\bottomrule
\end{tabular*}

\label{tab:gpu}
\end{table}

Each update records its wall-clock time on the training GPU after device synchronization, covering rollout generation, verifier execution, the backward pass, the optimizer step, and online-bank admission; held-out evaluation, checkpointing, and model loading are excluded.
The cumulative update time of each witness run converts its seed's matching cost into GPU time, linearly within an update, and Less GPU time divides the baseline's mean training time for its full budget by the mean of these times.
Initial bank verification is charged before the witness's first update, and building the self-bootstrapped bank is charged in full to every seed.
In the rows of Table~\ref{tab:gpu} marked $^\dagger$, the witness runs and the GRPO runs they pair with at 400 calls kept only their total wall-clock time, so their training time is that total minus evaluation.
At each call count, a witness run is charged its training time times the largest mean share that witness runs spend up to that count among settings timed per update.
A GRPO run is charged its training time reduced by the largest mean share that GRPO runs spend outside updates in those settings (\GPUOverhead\%).
Rebuilt without each of the \EnvCheckN{} settings timed per update, this rule recovers \EnvCheckMin\% to \EnvCheckMax\% of its measured Less GPU time.
For GRPO continued to 1,200 calls on \mbox{MBPP+}, training time is three times the 400-call training time of the six continued seeds.
Each seed's witness and baseline runs use the same GPU type, and every bootstrap replicate recomputes the target, matching costs, witness times, and baseline mean.
With \mbox{Qwen3-1.7B}, the interval of calls to match LUFFY-style guidance reaches zero on its \LuffyGPUPairs{} same-GPU seeds, so Table~\ref{tab:he} gives its Less GPU time as a lower bound, while its Fewer calls over all 12 seeds has a finite interval.

For larger-group GRPO ($G=8$), evaluations follow the first update reaching each common-grid call count, and the curve is interpolated onto that grid for AUC.
Final accuracy uses the observed endpoint.

Evaluation at $T=0.2$ decodes the final policies with the held-out bugs, sample count, prompt, and length caps used at $T=1$.

\FloatBarrier
\section{Additional Outcome-Reward, Ablation, and Witness-Choice Comparisons}
\label{app:effects}
REINFORCE in Tables~\ref{tab:algo} and~\ref{tab:ceonly} keeps only the policy-gradient term of Equation~(\ref{eq:loss}), and the random farther witness in Table~\ref{tab:choice} keeps both terms and changes only the target.
In Table~\ref{tab:ceonly}, on the 12 seeds of Table~\ref{tab:he}, CE-only online witnesses train with cross-entropy on their targets alone, and likelihood witnesses keep both terms but give each failure the bank member with the highest mean target-token log-probability under the current policy.
On \mbox{HumanEval+}, REINFORCE sees uniform rewards in \ReinforceZeroSmall\% and \ReinforceZeroLarge\% of its two-proposal groups with \mbox{Qwen3-1.7B} and \mbox{Qwen3-14B}.
Likelihood retrieval keeps the nearest witness for \LikNearSmall\% and \LikNearLarge\% of failures with a canonical form, and its targets lie on average \LikExtraSmall{} and \LikExtraLarge~canonical token edits beyond the nearest distance.

\begin{table}[!htb]
\caption{Repair witnesses outperform outcome-reward baselines on Algo-8. Accuracies are percentages: Base before training, and Witness and Baseline at 400 calls. With \mbox{Qwen3.5-4B}, REINFORCE pairs with its own online witness runs on the same 12 seeds. $\Delta$ values are witness minus baseline in points, and Fewer calls, Less GPU time, $>$, and dashes follow Section~\ref{sec:metrics}. Intervals are 95\% paired seed-bootstrap intervals over 12 seeds.}
\centering\small\setlength{\tabcolsep}{5pt}
\begin{tabular*}{\linewidth}{@{\extracolsep{\fill}}llrrrr@{\extracolsep{0pt}\hspace{3pt}}>{\scriptsize}l@{\extracolsep{\fill}\hspace{2\tabcolsep}}r@{\extracolsep{0pt}\hspace{3pt}}>{\scriptsize}l@{\extracolsep{\fill}\hspace{2\tabcolsep}}r@{\extracolsep{0pt}\hspace{3pt}}>{\scriptsize}l@{\extracolsep{\fill}\hspace{2\tabcolsep}}r@{\extracolsep{0pt}\hspace{3pt}}>{\scriptsize}l@{}}
\toprule
 &  &  &  &  & \multicolumn{2}{c}{$\Delta$AUC} & \multicolumn{2}{c}{$\Delta$ accuracy} & \multicolumn{2}{c}{Fewer} & \multicolumn{2}{c}{Less GPU} \\
Model & Method & Base & Witness & Baseline & \multicolumn{2}{c}{[95\% CI]} & \multicolumn{2}{c}{[95\% CI]} & \multicolumn{2}{c}{calls ($\times$)} & \multicolumn{2}{c}{time ($\times$)} \\
\midrule
\mbox{Qwen3-1.7B} & GRPO & 59.2 & 91.7 & 72.5 & +12.6 & \mbox{[8.8, 16.5]} & +19.2 & \mbox{[10.3, 30.1]} & 12.0 & \mbox{[3.5, 30.9]} & -- &  \\
 & REINFORCE & 59.2 & 91.7 & 63.4 & +12.3 & \mbox{[9.0, 15.6]} & +28.3 & \mbox{[17.5, 42.1]} & 19.9 & \mbox{[8.0, 61.0]} & -- &  \\
\mbox{SmolLM3-3B} & GRPO & 69.6 & 90.5 & 73.7 & +9.5 & \mbox{[7.4, 12.2]} & +16.8 & \mbox{[6.1, 32.4]} & $>$2.7 &  & $>$2.5 &  \\
\mbox{Granite-4.2-3B} & GRPO & 77.4 & 89.4 & 56.5 & +21.9 & \mbox{[16.8, 27.5]} & +32.9 & \mbox{[19.8, 48.4]} & 26.5 & \mbox{[6.0, 570.1]} & -- &  \\
\mbox{Qwen3.5-4B} & GRPO & 83.2 & 91.6 & 84.0 & +4.1 & \mbox{[1.2, 7.3]} & +7.6 & \mbox{[3.8, 11.7]} & 3.1 & \mbox{[1.8, 6.6]} & 2.7 & \mbox{[1.6, 5.6]} \\
 & REINFORCE & 83.2 & 91.9 & 59.3 & +13.9 & \mbox{[8.2, 19.7]} & +32.6 & \mbox{[14.7, 52.6]} & $>$8.0 &  & -- &  \\
\bottomrule
\end{tabular*}

\label{tab:algo}
\end{table}
\begin{table}[!htb]
\caption{Cross-entropy on witness targets carries the lead of online witnesses over GRPO on \mbox{HumanEval+}: CE-only keeps most of it, REINFORCE matches GRPO, and likelihood witnesses learn faster still. Accuracies are at 400 calls, and $\Delta$ is online witness minus the row's method in points with 95\% paired seed-bootstrap intervals over 12 seeds. CE-only lead kept is CE-only minus GRPO as a percentage of online witness minus GRPO.}
\centering\small
\begin{tabular*}{\linewidth}{@{\extracolsep{\fill}}llrrrr@{\extracolsep{0pt}\hspace{3pt}}>{\scriptsize}l@{\extracolsep{\fill}\hspace{2\tabcolsep}}r@{\extracolsep{0pt}\hspace{3pt}}>{\scriptsize}l@{\extracolsep{\fill}\hspace{2\tabcolsep}}r@{\extracolsep{0pt}\hspace{3pt}}>{\scriptsize}l@{}}
\toprule
 &  &  & \multicolumn{2}{c}{Accuracy (\%)} & \multicolumn{2}{c}{} & \multicolumn{4}{c}{$\Delta$ accuracy [95\% CI]} \\
\cmidrule(lr){4-5}\cmidrule(lr){8-11}
Model & Method & AUC (\%) & $T{=}1$ & $T{=}0.2$ & \multicolumn{2}{c}{$\Delta$AUC [95\% CI]} & \multicolumn{2}{c}{$T{=}1$} & \multicolumn{2}{c}{$T{=}0.2$} \\
\midrule
\mbox{Qwen3-1.7B} & Online witness & 32.5 & 58.6 & 73.2 & -- &  & -- &  & -- &  \\
 & CE-only & 32.1 & 57.9 & 72.5 & +0.4 & \mbox{[$-$2.3, 2.8]} & +0.7 & \mbox{[$-$3.6, 4.7]} & +0.7 & \mbox{[$-$2.9, 4.0]} \\
 & REINFORCE & 13.2 & 16.9 & 17.2 & +19.3 & \mbox{[15.6, 22.6]} & +41.7 & \mbox{[34.4, 49.2]} & +56.0 & \mbox{[47.3, 64.4]} \\
 & Likelihood witness & 40.5 & 62.6 & 73.4 & $-$8.0 & \mbox{[$-$10.1, $-$6.0]} & $-$4.0 & \mbox{[$-$8.0, $-$0.2]} & $-$0.2 & \mbox{[$-$3.8, 3.6]} \\
 & GRPO & 14.9 & 15.6 & 15.0 & +17.6 & \mbox{[14.8, 20.4]} & +43.0 & \mbox{[32.7, 51.0]} & +58.2 & \mbox{[50.8, 64.3]} \\
 & CE-only lead kept (\%) & 97.9 & 98.3 & 98.7 &  &  &  &  &  &  \\
\addlinespace[3pt]
\mbox{Qwen3-14B} & Online witness & 52.4 & 75.5 & 81.9 & -- &  & -- &  & -- &  \\
 & CE-only & 50.1 & 75.1 & 83.8 & +2.3 & \mbox{[0.3, 4.3]} & +0.4 & \mbox{[$-$4.9, 5.3]} & $-$1.9 & \mbox{[$-$4.1, 0.4]} \\
 & REINFORCE & 29.1 & 32.4 & 33.7 & +23.3 & \mbox{[18.0, 28.2]} & +43.1 & \mbox{[34.3, 52.0]} & +48.2 & \mbox{[38.7, 57.8]} \\
 & Likelihood witness & 55.5 & 77.8 & 85.1 & $-$3.1 & \mbox{[$-$4.7, $-$1.5]} & $-$2.3 & \mbox{[$-$6.8, 2.4]} & $-$3.2 & \mbox{[$-$6.5, 0.2]} \\
 & GRPO & 29.2 & 38.7 & 38.6 & +23.2 & \mbox{[18.7, 28.4]} & +36.8 & \mbox{[28.9, 45.6]} & +43.3 & \mbox{[36.7, 51.3]} \\
 & CE-only lead kept (\%) & 90.2 & 98.9 & 104.5 &  &  &  &  &  &  \\
\bottomrule
\end{tabular*}

\label{tab:ceonly}
\end{table}
\begin{table}[!htb]
\caption{By AUC over the 400-call budget, the nearest witness learns faster than a random farther witness in all 14 comparisons. Entries are nearest minus random farther witness in AUC points, with 95\% paired seed-bootstrap intervals over 12 seeds (53 for Algo-8 with \mbox{Qwen3-1.7B}).}
\centering\small
\begin{tabular*}{\linewidth}{@{\extracolsep{\fill}}lr@{\extracolsep{0pt}\hspace{3pt}}>{\scriptsize}l@{\extracolsep{\fill}\hspace{2\tabcolsep}}r@{\extracolsep{0pt}\hspace{3pt}}>{\scriptsize}l@{\extracolsep{\fill}\hspace{2\tabcolsep}}r@{\extracolsep{0pt}\hspace{3pt}}>{\scriptsize}l@{}}
\toprule
Model & \multicolumn{2}{c}{Algo-8} & \multicolumn{2}{c}{Hanoi-Hard} & \multicolumn{2}{c}{\mbox{HumanEval+}} \\
\midrule
\mbox{Qwen3-1.7B} & +7.6 & \mbox{[6.6, 8.7]} & +4.5 & \mbox{[2.0, 7.2]} & +4.3 & \mbox{[2.3, 6.1]} \\
\mbox{SmolLM3-3B} & +5.5 & \mbox{[3.6, 7.8]} & +2.0 & \mbox{[$-$0.1, 4.0]} & +0.3 & \mbox{[$-$1.6, 2.1]} \\
\mbox{Granite-4.2-3B} & +4.4 & \mbox{[1.1, 8.2]} & -- &  & -- &  \\
\mbox{Qwen3.5-4B} & +1.4 & \mbox{[$-$1.1, 3.8]} & -- &  & -- &  \\
\mbox{Qwen3-14B} & -- &  & +7.9 & \mbox{[5.8, 10.1]} & +2.5 & \mbox{[0.4, 4.3]} \\
\mbox{Qwen3-14B}, self-bootstrapped bank & -- &  & +6.8 & \mbox{[4.7, 8.8]} & -- &  \\
\mbox{Qwen3.6-27B} & -- &  & +5.5 & \mbox{[2.4, 8.5]} & -- &  \\
\mbox{Qwen3-32B} & -- &  & +4.3 & \mbox{[2.0, 6.8]} & +2.5 & \mbox{[$-$1.0, 6.3]} \\
\addlinespace[3pt]
All 14 comparisons, pooled & \multicolumn{6}{c}{+4.3\hspace{3pt}{\scriptsize \mbox{[3.6, 4.9]}}} \\
\bottomrule
\end{tabular*}

\label{tab:choice}
\end{table}

\FloatBarrier
\section{Self-Bootstrapped Banks and Cold-Start Repair}
\label{app:selfbank}
On Hanoi-Hard with \mbox{Qwen3-14B}, witnesses on the self-bootstrapped bank reach \SelfHanoiAcc\% \SelfHanoiAccCI{} after 230 total calls, the 102 construction calls included, and match the 1,600-call accuracy of GRPO, \SelfHanoiGRPO\% \SelfHanoiGRPOCI, with 9.0~times fewer total calls \mbox{[6.0, 9.6]}.
Against the curated bank, they gain 10.9 AUC points \mbox{[9.1, 12.5]}.
On 48 fixed failed proposals for 22 training defects, base \mbox{Qwen3-14B} gives the nearest self-bootstrapped witness a per-token negative log-likelihood 0.23~nats \mbox{[0.16, 0.30]} lower than the nearest curated witness, with defects resampled in a paired bootstrap.

\begin{table}[!htb]
\caption{With a self-bootstrapped bank and no reference implementation, online witnesses outperform GRPO, RFT, and LUFFY-style guidance on \mbox{HumanEval+}. The bank starts from the model's own verified programs and admits passing proposals during training. Offline witnesses use the initial bank. At $T=1$, Outside bank is measured against the initial bank, and $\Delta$ accuracy and Fewer calls follow Table~\ref{tab:he} with 95\% paired seed-bootstrap intervals over 12 seeds.}
\centering\small
\begin{tabular*}{\linewidth}{@{\extracolsep{\fill}}llrrrrr@{\extracolsep{0pt}\hspace{3pt}}>{\scriptsize}l@{\extracolsep{\fill}\hspace{2\tabcolsep}}r@{\extracolsep{0pt}\hspace{3pt}}>{\scriptsize}l@{}}
\toprule
 &  &  & \multicolumn{2}{c}{Accuracy (\%)} &  & \multicolumn{2}{c}{} & \multicolumn{2}{c}{} \\
\cmidrule(lr){4-5}
Model & Method & Calls & $T{=}1$ & $T{=}0.2$ & Outside bank (\%) & \multicolumn{2}{c}{$\Delta$ accuracy [95\% CI]} & \multicolumn{2}{c}{Fewer calls ($\times$)} \\
\midrule
\mbox{Qwen3-1.7B} & Online witness & 400 & 60.5 & 69.2 & 51.3 & -- &  & -- &  \\
 & Offline witness & 0 & 55.5 & 65.4 & 33.1 & -- &  & -- &  \\
 & GRPO & 400 & 15.6 & 15.0 & 15.2 & +44.9 & \mbox{[36.1, 51.8]} & 12.0 & \mbox{[4.4, 31.1]} \\
 & RFT & 400 & 24.7 & 24.6 & 23.9 & +35.8 & \mbox{[26.1, 44.7]} & 4.1 & \mbox{[2.8, 11.5]} \\
 & LUFFY-style & 400 & 3.8 & 9.1 & 3.2 & +56.7 & \mbox{[51.6, 61.4]} & 40.5 & \mbox{[24.9, 129.1]} \\
\addlinespace[3pt]
\mbox{Qwen3-14B} & Online witness & 400 & 65.5 & 72.6 & 50.9 & -- &  & -- &  \\
 & Offline witness & 0 & 59.3 & 63.0 & 33.5 & -- &  & -- &  \\
 & GRPO & 400 & 38.7 & 38.6 & 34.6 & +26.8 & \mbox{[20.1, 36.1]} & 5.0 & \mbox{[3.6, 11.9]} \\
 & RFT & 400 & 45.6 & 47.1 & 42.7 & +19.9 & \mbox{[14.5, 25.8]} & 3.8 & \mbox{[2.4, 4.7]} \\
 & LUFFY-style & 400 & 37.1 & 44.5 & 34.7 & +28.4 & \mbox{[18.5, 39.6]} & 5.2 & \mbox{[3.2, 16.4]} \\
\bottomrule
\end{tabular*}

\label{tab:selfhe}
\end{table}

On the self-bootstrapped banks, each run admits 90.0 new programs \mbox{[84.2, 95.8]} with \mbox{Qwen3-1.7B} and 85.7 \mbox{[78.1, 94.0]} with \mbox{Qwen3-14B}, so the banks grow from 84 to 174 programs and from 54 to 140.

On the \SelfCurCovSmall{} of \SelfCurFnSmall{} functions for \mbox{Qwen3-1.7B} and \SelfCurCovLarge{} of \SelfCurFnLarge{} for \mbox{Qwen3-14B} that begin with a self-generated program (Table~\ref{tab:selfcold}), the self-bootstrapped bank held fixed during training leads the curated bank at $T=1$ by \SelfStaticPooled~points \SelfStaticPooledCI{} pooled.

\begin{table}[!htb]
\caption{Online witnesses learn from a cold start on \mbox{MBPP+} and LiveCodeBench and finish above GRPO, DAPO, and Dr.~GRPO on training and unseen functions alike. With a program and Empty bank split the training functions by initial bank at $T=1$: With a program covers 39 and \SmolColdCovered{} \mbox{MBPP+} functions (\mbox{Qwen3-1.7B}, \mbox{SmolLM3-3B}) and 51 LiveCodeBench functions, and Empty bank the remaining 64, \SmolColdEmpty, and 109. Unseen functions are absent from training and the bank. Method $-$ GRPO is the row's accuracy minus that of GRPO at the same budget and $T=1$ in points, with 95\% paired seed-bootstrap intervals over 12 seeds (six for the 1,200-call block). $\Delta$AUC over the 400-call budget and Fewer calls compare online witnesses with GRPO on held-out bugs of the training functions, and $>$ and dashes follow Section~\ref{sec:metrics}.}
\centering\small\setlength{\tabcolsep}{3pt}
\begin{tabular*}{\linewidth}{@{\extracolsep{\fill}}lrrrrr@{\extracolsep{0pt}\hspace{3pt}}>{\scriptsize}l@{\extracolsep{\fill}\hspace{2\tabcolsep}}r@{\extracolsep{0pt}\hspace{3pt}}>{\scriptsize}l@{\extracolsep{\fill}\hspace{2\tabcolsep}}r@{\extracolsep{0pt}\hspace{3pt}}>{\scriptsize}l@{\extracolsep{\fill}\hspace{2\tabcolsep}}rrr@{\extracolsep{0pt}\hspace{3pt}}>{\scriptsize}l@{}}
\toprule
 & \multicolumn{10}{c}{Held-out bugs of the training functions} & \multicolumn{4}{c}{Unseen functions} \\
\cmidrule(lr){2-11}\cmidrule(lr){12-15}
 & \multicolumn{2}{c}{All} & With a & Empty & \multicolumn{2}{c}{Method} & \multicolumn{2}{c}{} & \multicolumn{2}{c}{Fewer} &  &  & \multicolumn{2}{c}{Method} \\
\cmidrule(lr){2-3}
Method & $T{=}1$ & $T{=}0.2$ & program & bank & \multicolumn{2}{c}{$-$ GRPO} & \multicolumn{2}{c}{$\Delta$AUC} & \multicolumn{2}{c}{calls ($\times$)} & $T{=}1$ & $T{=}0.2$ & \multicolumn{2}{c}{$-$ GRPO} \\
\midrule
\multicolumn{15}{@{}l}{\textit{\mbox{MBPP+}, \mbox{Qwen3-1.7B}}} \\
Online & 22.6 & 28.6 & 52.1 & 4.6 & +22.3 & \mbox{[20.8, 23.8]} & +13.1 & \mbox{[12.6, 13.5]} & $>$16 &  & 11.9 & 11.9 & +11.6 & \mbox{[8.7, 14.7]} \\
GRPO & 0.3 & 0.3 & 0.9 & 0.0 & -- &  & -- &  & -- &  & 0.3 & 0.3 & -- &  \\
DAPO & 0.5 & 0.7 & -- & -- & +0.2 & \mbox{[$-$0.4, 0.6]} & -- &  & -- &  & 0.3 & 0.3 & 0.0 & \mbox{[$-$1.0, 1.0]} \\
Dr.~GRPO & 0.1 & 0.1 & -- & -- & $-$0.2 & \mbox{[$-$0.6, 0.2]} & -- &  & -- &  & 0.0 & 0.0 & $-$0.3 & \mbox{[$-$1.0, 0.0]} \\
\addlinespace[3pt]
\multicolumn{15}{@{}l}{\textit{\mbox{MBPP+}, \mbox{SmolLM3-3B}}} \\
Online & 31.4 & 37.9 & 56.9 & 3.4 & +16.6 & \mbox{[12.0, 21.0]} & +14.0 & \mbox{[12.3, 15.6]} & 7.2 & \mbox{[5.4, 9.4]} & 22.1 & 24.1 & +10.3 & \mbox{[5.7, 14.9]} \\
GRPO & 14.8 & 15.0 & 26.8 & 1.7 & -- &  & -- &  & -- &  & 11.8 & 13.2 & -- &  \\
\addlinespace[3pt]
\multicolumn{15}{@{}l}{\textit{\mbox{MBPP+}, \mbox{Qwen3-1.7B}, 1,200 calls}} \\
Online & 35.9 & 39.0 & 78.2 & 10.2 & +33.0 & \mbox{[30.1, 35.4]} & -- &  & 29.4 & \mbox{[15.6, 85.8]} & 14.7 & 15.4 & +12.1 & \mbox{[5.1, 17.9]} \\
GRPO & 2.9 & 2.9 & 7.3 & 0.3 & -- &  & -- &  & -- &  & 2.6 & 1.3 & -- &  \\
\addlinespace[3pt]
\multicolumn{15}{@{}l}{\textit{LiveCodeBench, \mbox{Qwen3-8B}}} \\
Online & 15.8 & 17.6 & 47.1 & 1.2 & +14.6 & \mbox{[13.2, 15.8]} & +8.2 & \mbox{[7.7, 8.8]} & 18.0 & \mbox{[9.1, 43.3]} & 10.6 & 11.0 & +8.1 & \mbox{[6.4, 9.7]} \\
GRPO & 1.2 & 0.8 & 3.8 & 0.1 & -- &  & -- &  & -- &  & 2.5 & 1.9 & -- &  \\
\bottomrule
\end{tabular*}

\label{tab:cold}
\end{table}

GRPO receives no reward gradient in \ColdZeroMbpp\% of updates on \mbox{MBPP+} with \mbox{Qwen3-1.7B}, \ColdZeroSmol\% with \mbox{SmolLM3-3B}, and \ColdZeroLcb\% on LiveCodeBench, and DAPO's dynamic sampling discards 99.0\% of its \mbox{MBPP+} groups, so a run averages one update and 5 of 12 never update.

\FloatBarrier
\Needspace{8\baselineskip}
\section{The Witness Family and Outside-Bank Correctness}
\label{app:copy}
On \mbox{HumanEval+}, success-projected witnesses gain \ProjAucSmall~AUC points \ProjAucSmallCI{} over online witnesses with \mbox{Qwen3-1.7B}, \ProjAucSmol{} \ProjAucSmolCI{} with \mbox{SmolLM3-3B}, and \ProjAucLarge{} \ProjAucLargeCI{} with \mbox{Qwen3-14B}.
Pooled over the three models, they finish \ProjPooledOnline~points \ProjPooledOnlineCI{} above online witnesses and \ProjPooledOffline~points \ProjPooledOfflineCI{} above offline witnesses at $T=1$.

\begin{table}[!htb]
\caption{Online witnesses finish above offline witnesses in all ten settings and write more correct programs outside the bank in all eight settings that count them. Accuracies are final percentages at $T=1$ on held-out defects (of the training functions for \mbox{MBPP+} and LiveCodeBench), and $\Delta$ is online minus offline witness in points with 95\% paired seed-bootstrap intervals over 12 seeds. Outside-bank correctness counts against the initial bank on self-bootstrapped banks, and the Hanoi-Hard rows omit it because their held-out defects are mutations of bank members. The Self-bootstrapped banks row pools the four self-bootstrapped settings.}
\centering\small
\begin{tabular*}{\linewidth}{@{\extracolsep{\fill}}llrrr@{\extracolsep{0pt}\hspace{3pt}}>{\scriptsize}l@{\extracolsep{\fill}\hspace{2\tabcolsep}}r@{\extracolsep{0pt}\hspace{3pt}}>{\scriptsize}l@{}}
\toprule
 &  &  &  & \multicolumn{2}{c}{$\Delta$ accuracy} & \multicolumn{2}{c}{$\Delta$ outside bank} \\
Task & Model & Online (\%) & Offline (\%) & \multicolumn{2}{c}{[95\% CI]} & \multicolumn{2}{c}{[95\% CI]} \\
\midrule
\mbox{HumanEval+} & \mbox{Qwen3-1.7B} & 58.6 & 57.4 & +1.2 & \mbox{[$-$1.9, 4.6]} & +9.8 & \mbox{[7.3, 12.3]} \\
 & \mbox{SmolLM3-3B} & 66.1 & 64.1 & +2.0 & \mbox{[$-$2.2, 6.7]} & +19.0 & \mbox{[14.9, 23.0]} \\
 & \mbox{Qwen3-14B} & 75.5 & 74.4 & +1.1 & \mbox{[$-$4.2, 6.5]} & +17.7 & \mbox{[11.7, 23.5]} \\
 & \mbox{Qwen3-32B} & 74.2 & 69.1 & +5.1 & \mbox{[0.2, 9.8]} & +26.3 & \mbox{[21.4, 30.7]} \\
\addlinespace[3pt]
 & \mbox{Qwen3-1.7B}, self-bootstrapped bank & 60.5 & 55.5 & +5.0 & \mbox{[0.7, 9.0]} & +18.2 & \mbox{[13.6, 21.9]} \\
 & \mbox{Qwen3-14B}, self-bootstrapped bank & 65.5 & 59.3 & +6.2 & \mbox{[2.3, 10.0]} & +17.4 & \mbox{[11.8, 23.3]} \\
\addlinespace[3pt]
Hanoi-Hard & \mbox{Qwen3-1.7B} & 77.0 & 66.8 & +10.2 & \mbox{[5.1, 15.9]} & -- &  \\
 & \mbox{Qwen3-14B} & 87.8 & 87.3 & +0.5 & \mbox{[$-$3.4, 4.8]} & -- &  \\
\addlinespace[3pt]
\mbox{MBPP+} & \mbox{Qwen3-1.7B} & 22.6 & 19.2 & +3.4 & \mbox{[1.3, 5.7]} & +4.4 & \mbox{[2.3, 6.7]} \\
\addlinespace[3pt]
LiveCodeBench & \mbox{Qwen3-8B} & 15.8 & 12.4 & +3.4 & \mbox{[2.2, 4.8]} & +3.7 & \mbox{[2.2, 5.2]} \\
\addlinespace[3pt]
\rlap{All ten settings, pooled} &  &  &  & +3.8 & \mbox{[2.5, 5.1]} & -- &  \\
\rlap{Self-bootstrapped banks, pooled} &  &  &  & +4.5 & \mbox{[2.9, 6.0]} & +10.9 & \mbox{[9.0, 12.8]} \\
\bottomrule
\end{tabular*}

\label{tab:onoff}
\end{table}
\begin{table}[!htb]
\caption{Started from the same offline-witness policy, online witnesses outperform GRPO on \mbox{HumanEval+}. Offline $\to$ online and Offline $\to$ GRPO continue each seed's final offline-witness policy with 400 calls of online witnesses or GRPO, and Online witness and Offline witness are the runs of Table~\ref{tab:he}. Accuracies are final percentages, and differences are in points with 95\% paired seed-bootstrap intervals over the 12 seeds of Table~\ref{tab:he} with \mbox{Qwen3-14B} and \SftPairsSmall{} of them with \mbox{Qwen3-1.7B}, whose Online witness and Offline witness rows therefore differ from Table~\ref{tab:he}.}
\centering\small\setlength{\tabcolsep}{3pt}
\begin{tabular*}{\linewidth}{@{\extracolsep{\fill}}llrrr@{\extracolsep{0pt}\hspace{3pt}}>{\scriptsize}l@{\extracolsep{\fill}\hspace{2\tabcolsep}}r@{\extracolsep{0pt}\hspace{3pt}}>{\scriptsize}l@{\extracolsep{\fill}\hspace{2\tabcolsep}}r@{\extracolsep{0pt}\hspace{3pt}}>{\scriptsize}l@{\extracolsep{\fill}\hspace{2\tabcolsep}}r@{\extracolsep{0pt}\hspace{3pt}}>{\scriptsize}l@{\extracolsep{\fill}\hspace{2\tabcolsep}}r@{\extracolsep{0pt}\hspace{3pt}}>{\scriptsize}l@{}}
\toprule
 &  & \multicolumn{2}{c}{Accuracy (\%)} & \multicolumn{6}{c}{Offline $\to$ online $-$ row [95\% CI]} & \multicolumn{4}{c}{Offline $\to$ GRPO $-$ row [95\% CI]} \\
\cmidrule(lr){3-4}\cmidrule(lr){5-10}\cmidrule(lr){11-14}
Model & Method & $T{=}1$ & $T{=}0.2$ & \multicolumn{2}{c}{$\Delta$AUC} & \multicolumn{2}{c}{$T{=}1$} & \multicolumn{2}{c}{$T{=}0.2$} & \multicolumn{2}{c}{$T{=}1$} & \multicolumn{2}{c}{$T{=}0.2$} \\
\midrule
\mbox{Qwen3-1.7B} & Offline $\to$ online & 72.7 & 82.3 & -- &  & -- &  & -- &  & -- &  & -- &  \\
 & Offline $\to$ GRPO & 62.3 & 69.4 & +4.1 & \mbox{[2.5, 5.6]} & +10.4 & \mbox{[3.0, 18.0]} & +12.9 & \mbox{[8.1, 18.6]} & -- &  & -- &  \\
 & Online witness & 57.8 & 73.9 & -- &  & +14.9 & \mbox{[9.2, 20.6]} & +8.4 & \mbox{[4.8, 12.2]} & +4.5 & \mbox{[0.0, 8.6]} & $-$4.5 & \mbox{[$-$8.9, $-$0.2]} \\
 & Offline witness & 55.9 & 69.5 & -- &  & +16.8 & \mbox{[12.3, 21.4]} & +12.8 & \mbox{[7.7, 18.4]} & +6.4 & \mbox{[1.6, 10.0]} & $-$0.1 & \mbox{[$-$7.0, 4.8]} \\
\addlinespace[3pt]
\mbox{Qwen3-14B} & Offline $\to$ online & 84.1 & 91.4 & -- &  & -- &  & -- &  & -- &  & -- &  \\
 & Offline $\to$ GRPO & 75.6 & 79.4 & +3.6 & \mbox{[1.4, 5.9]} & +8.5 & \mbox{[1.9, 15.5]} & +12.0 & \mbox{[6.4, 17.1]} & -- &  & -- &  \\
 & Online witness & 75.5 & 81.9 & -- &  & +8.6 & \mbox{[2.6, 14.8]} & +9.5 & \mbox{[4.1, 13.4]} & +0.1 & \mbox{[$-$3.7, 3.9]} & $-$2.5 & \mbox{[$-$6.5, 1.5]} \\
 & Offline witness & 74.4 & 82.5 & -- &  & +9.7 & \mbox{[5.0, 14.5]} & +8.9 & \mbox{[4.1, 13.0]} & +1.2 & \mbox{[$-$3.6, 5.9]} & $-$3.1 & \mbox{[$-$7.5, 1.0]} \\
\bottomrule
\end{tabular*}

\label{tab:sftrl}
\end{table}

In Table~\ref{tab:sftrl}, each seed's offline-witness adapter is merged into the base weights, and GRPO or online witnesses train a new adapter on the merged policy for 400 calls, with the merged policy as the KL reference of GRPO.
From this shared start, online witnesses reach the 400-call accuracy of GRPO after \SftCallsSmall{} calls with \mbox{Qwen3-1.7B} and \SftCallsLarge{} with \mbox{Qwen3-14B}, \SftFoldSmall~times fewer \SftFoldSmallCI{} and \SftFoldLarge~times fewer \SftFoldLargeCI.
At $T=0.2$, GRPO ends no higher than the offline-witness policy it starts from or than online witnesses trained from the base model with the same 400 calls.

\FloatBarrier

\end{document}